\def\runningtitle{Role support in knowledge-graph error ranking}
\documentclass[11pt,a4paper]{article}
\usepackage[margin=25mm]{geometry}
\usepackage[T1]{fontenc}
\usepackage{lmodern,microtype,amsmath,amssymb,amsthm,booktabs,tabularx,array,graphicx,xcolor,enumitem,fancyhdr}
\usepackage[hidelinks]{hyperref}
\definecolor{ink}{HTML}{17324D}
\definecolor{muted}{HTML}{526475}
\definecolor{wash}{HTML}{EDF3F7}
\newcommand{\E}{\mathbb{E}}
\newcommand{\Prb}{\mathbb{P}}

\newcommand{\ind}{\mathbf{1}}
\newcommand{\auc}{\operatorname{AUC}}
\newcommand{\tv}{\operatorname{TV}}
\newtheorem{proposition}{Proposition}
\newtheorem{corollary}{Corollary}
\theoremstyle{definition}\newtheorem{example}{Example}
\setlist{nosep,leftmargin=1.5em}
\newcommand{\sourceurl}[1]{\url{#1}}
\title{Role Support in Knowledge-Graph Error Ranking:\\Predictor Regimes and Evaluation Policies}
\author{Chorok Lee\\Korea Advanced Institute of Science and Technology}
\date{}
\begin{document}
\maketitle
\begin{abstract}
When do role-specific training counts improve knowledge-graph error ranking? We evaluate fifteen from-scratch predictors, nine development-selected continuation stages, and six fixed ensembles across three static graphs. A positive penalty for missing training support has no stable error-ranking orientation: stronger ComplEx training reverses its pooled error AUROC from 0.689 to 0.364 on FB15k-237, and from 0.729 to 0.430 in a prospectively specified CoDEx-S check. Pair decomposition shows that reversal also survives pair-weighted within-relation evaluation, but not equal-relation averaging. The statement depends on which relations and query pairs receive weight. Learned continuous counts remain useful: they improve AURC and Brier beyond explicit ensemble disagreement in twelve original ensemble/head comparisons and satisfy all four predefined external-dataset AURC contrasts. Binary indicators add no information beyond their defining counts and have less stable fitting effects. Candidate, target, population and acceptance policies change the comparisons; fixed development thresholds do not preserve a common test answer rate. The external study was fixed before inspection of its test outcomes, whereas subsequent explanations and reused-graph controls are explicitly exploratory. These measurements support relation-aware count baselines and policy-specific support diagnostics, not a new uncertainty decomposition, causal training mechanism, calibrated factual truth, or graph-population guarantee. Saved checkpoints, ledgers and separately authored numerical implementation replays make the scoped findings inspectable.
\end{abstract}

\section{Introduction}
Knowledge-graph completion ranks possible answers to a query; a second score can rank those predictions for acceptance or scrutiny. Relation-specific training counts are natural inputs because they record how often an entity has appeared in a relation's head or tail role. Their nonzero sets are established observed domain/range constructions~\cite{typeconstraints}. Directional peer-count and tail--relation-degree associations with link prediction are also established~\cite{rossipeers,degreebias}. In an incomplete graph, however, training absence does not establish falsehood, and the meaning of support can change with the predictor producing the answer.

We ask whether role support has a stable error-ranking interpretation, and whether learned counts remain useful after strengthening the comparator. The earlier version of this preprint proposed a positive penalty for missing support. We test that specific heuristic; we do not attribute it to uncertainty methods generally. A missing-role bit also adds no information beyond its defining count, although supplying it explicitly can change a restricted fitted model.

Three questions organize the evidence. First, does the positive missing-support ordering survive a more accurate predictor and alternative relation weightings? Second, do continuous counts improve error scoring beyond confidence, frequencies, explicit ensemble disagreement, and relation identity? Third, do those conclusions survive changes in population, recorded targets, candidates, and acceptance policy? We answer with fifteen from-scratch predictors, nine continuation stages, and six fixed ensembles across three graphs. CoDEx-S was chosen and its primary contrasts fixed before accessing its test outcomes; the earlier two-graph extensions and later explanations are exploratory.

The contribution is an empirical boundary finding with an executable decision-time information contract. Raw support can reverse its ordering, while learned counts retain a measured, often small, increment against stronger comparators. These are conditional results for recorded-answer recovery, rather than a new uncertainty decomposition or a causal account of training. The underlying role sets, predictor recipes, ensembles, and decision principles are established. Complete methods and unfavorable comparisons are retained in the appendices; Appendix~\ref{app:history} separately reconciles unreproduced historical claims from the earlier preprint~\cite{original}.

\section{Task, information, and supporting principles}\label{sec:contract}
\paragraph{The prediction and error must be fixed first.}
For a unique tail query $(h,r,?)$, a frozen predictor chooses $\widehat t$ from training-seen entities after masking only training-known answers. Its error is $L=\ind\{\widehat t\notin T_{h,r}\}$, where $T_{h,r}$ is the recorded held-out target set. Queries receive equal weight. An unrecorded true answer can count as an error: this is observed-target recovery, not factual-truth classification or standard filtered Hits@1. Paired scorer comparisons keep queries, predictions, targets, and candidates identical.

All decision features are computed before revealing the held-out targets. Define directional training counts $n_H(e,r)$ and $n_T(e,r)$, and missing-role bits $b_H=\ind\{n_H(h,r)=0\}$, $b_T=\ind\{n_T(\widehat t,r)=0\}$. The raw score $U_C=(b_H+b_T)/2$ gives equal positive penalties to both directions. Learned heads instead receive separate bits or $\log(1+n_H),\log(1+n_T)$. Features use the \emph{predicted} tail: a training-only count evaluated at the hidden true tail can still leak the answer (Appendix~\ref{app:diagnostics}). Reciprocal augmentation never changes the original counts.

\paragraph{Ranking and probability scoring answer different questions.}
At requested batch coverage $q$, retain $k=\lfloor qn\rfloor$ lowest-score queries and measure their error fraction. Tied boundaries are averaged analytically under uniform random selection. AURC is the discrete mean of this expected risk over ranks $k=1,\ldots,n$; lower is better. Error AUROC awards half credit to tied error--correct pairs; higher is better. Brier score evaluates a fitted error probability by mean squared error. We report risk at 50\% and 85\% coverage and the same-prediction hindsight floor. A threshold selected on development data is a separate policy whose test acceptance can change. Detailed conventions appear in Appendix~\ref{app:contract}.

\paragraph{What the mathematics establishes.}
Appendix~\ref{app:theory} supplies proofs and explicit assumptions. For fixed endpoint-feature distributions $Q_0,Q_1$ with total variation $d$, any score of those features obeys the sharp bound $|\mathrm{AUROC}-\tfrac12|\leq d-d^2/2$. This novelty-discrimination statement does not automatically apply to relation-dependent predicted endpoints. For a fixed error outcome and nested information, the Bayes Brier improvement is exactly
\[
 \E[(L-\eta_Z)^2]-\E[(L-\eta_{ZR})^2]
 =\E[(\eta_{ZR}-\eta_Z)^2],
 \qquad \eta_Z=\E[L\mid Z],\quad\eta_{ZR}=\E[L\mid Z,R].
\]
Extra context can therefore help an oracle precisely when it changes conditional error probabilities. At equal population coverage $q$, selecting the lowest true error probabilities is optimal. Writing $\eta$ for these conditional error probabilities, replacing them by a fixed estimate $s\in[0,1]$ gives excess selective risk at most
\[
 \min\!\left\{1,\frac{\sqrt{2\min(q,1-q)\,v}}q\right\},
 \qquad v=\E[(s-\eta)^2].
\]
These standard projection and plug-in arguments explain the target of learning; they do not certify our fitted heads, establish their approximation error, or guarantee threshold transfer. Calibration alone is insufficient for useful ranking. Finally, $b_a=\ind\{n_a=0\}$ implies that bits add no conditional information beyond exact counts; any fitted benefit concerns representation or estimation.

\section{Empirical findings}\label{sec:experiments}
We evaluate role support through three questions: whether a fixed missing-support penalty preserves its orientation, whether learned counts add to strong specified comparators, and whether the result survives a change in evaluation policy. The evidence comprises three static graphs, three optimizer seeds ($17,29,43$), and the combinations in Table~\ref{tab:main-experiment-matrix}. FB15k-237 and WN18RR informed the hypotheses and subsequent controls. CoDEx-S~\cite{codex} was selected from published metadata before fetching its triples; its training and primary comparison plan preceded data access, and all six predictor states and fitted heads preceded the first parsing of test outcomes. This external check is prospectively specified, not a public preregistration or independent execution. Its relation-pair decomposition was added after test inspection.

Every outcome is \emph{setwise observed-target recovery}: one query $(h,r)$ receives one prediction and equal weight, and is correct if that prediction belongs to its recorded held-out answer set. Training-known targets and unseen training identifiers are excluded; primary test keys overlapping validation are removed. Candidates are training entities excluding training-known answers, with no held-out filtering. Counts use training triples and features use the predicted tail. Thus the task measures recovery of recorded answers, including multiple answers per query, rather than factual falsehood or standard filtered link-ranking performance. Full protocols and attrition appear in Section~\ref{sec:new-experiments} and Table~\ref{tab:attrition}.

Higher scores rank expected errors higher. We report tie-aware error AUROC, discrete AURC averaging selective risk over every retained count, Brier loss for fitted error probabilities, and risks at 50\% and 85\% batch coverage. Tied score blocks are averaged analytically, independently of correctness. The original recorded contrast was binary-role LR minus Base LR at 85\%; later comparisons remain separately identified. All specified heads and adverse outcomes are retained. Seed means describe optimization variability on shared query populations, and their ranges are not confidence intervals or graph replications.

\begin{table}[t]
\centering\footnotesize
\setlength{\tabcolsep}{4pt}
\begin{tabularx}{\linewidth}{@{}>{\raggedright\arraybackslash}p{0.18\linewidth}>{\raggedright\arraybackslash}p{0.17\linewidth}>{\raggedright\arraybackslash}X>{\raggedright\arraybackslash}X@{}}\toprule
Population & Test queries & Predictor and comparisons & Evidence status\\\midrule
CoDEx-S primary & 1,105 & Three sampled ComplEx predictors and three reciprocal stages; counts, matched relation-ID tuning, thresholds & Pretest primary plan and supplemental relation-ID plan; pair explanation post-result\\
FB15k-237 / WN18RR primary & 10,301 / 2,689 & Six reciprocal stages, six fixed three-member ensembles; count, disagreement and relation-ID controls & Reused graphs; exploratory extensions\\
FB15k-237 / WN18RR original & 10,301 / 2,689 & Twelve sampled DistMult/ComplEx predictors; representation controls; DistMult candidate and threshold changes & Original contrast recorded before scoring; later controls exploratory\\
Grouped held-out pools & Original: 11,217 / 2,664; reciprocal: 5,568 / 1,368 & Original twelve and selected reciprocal six; refitted heads, target-set sensitivity & Reused outcomes; reciprocal checkpoint-selection bin excluded\\\bottomrule
\end{tabularx}
\caption{Executed comparisons, not independent replications. Counts of predictors include repeated seeds on the same graph; reciprocal stages reuse their sampled parent. Each ensemble is fixed once. The two grouped designs have different eligible bins and denominators.}
\label{tab:main-experiment-matrix}
\end{table}

\subsection{When does a positive missing-role penalty reverse?}
The more accurate regime is a warm-start reciprocal ComplEx stage: 32 complex components, eight additional full-softmax/N3 epochs, and development-only selection among epochs $0,2,4,6,8$~\cite{lacroix}. Its parents use 30 fixed epochs with sampled corruptions. Across the three graphs, mean recovery rises to 44--46\%, while the raw scalar $U_C=(b_H+b_T)/2$ reverses its pooled error-ranking direction on FB15k-237 and CoDEx-S (Table~\ref{tab:main-regime-auroc}). Every reciprocal seed on those graphs has AUROC below $1/2$. WN18RR remains above $1/2$. These are comparisons between fixed prediction regimes: the objective, optimizer, reciprocal parameters and budget change together, so improved accuracy is not identified as the cause of reversal.

\begin{table}[t]
\centering\small
\setlength{\tabcolsep}{4pt}
\begin{tabular}{@{}llrrrr@{}}\toprule
Graph & ComplEx stage & Accuracy (\%) & Pooled & Within & Macro\\\midrule
CoDEx-S & Reciprocal & 45.58 & 0.4297 & 0.4833 & 0.5645\\
 & Sampled & 14.36 & 0.7290 & 0.7039 & 0.8196\\
FB15k-237 & Reciprocal & 44.35 & 0.3639 & 0.4554 & 0.5508\\
 & Sampled & 10.05 & 0.6893 & 0.7813 & 0.7440\\
WN18RR & Reciprocal & 44.95 & 0.6341 & 0.5300 & 0.5932\\
 & Sampled & 37.47 & 0.7680 & 0.6288 & 0.7185\\\bottomrule
\end{tabular}
\caption{Raw missing-role error AUROC, averaged equally over three seeds. Within weights same-relation error--correct pairs equally; Macro weights eligible relations equally, omitting relations without both outcomes. Accuracy is setwise observed-target recovery. Pair decomposition is post-result explanation, including on CoDEx-S; complete masses and eligible-relation counts are in Tables~\ref{tab:relation-pair-diagnostic} and~\ref{tab:relation-pair-external}.}
\label{tab:main-regime-auroc}
\end{table}

The score merges two directions with different associations. After reciprocal training on FB15k-237, missing predicted-tail support falls from 55.50\% to 8.54\%. Among supported-tail predictions, mean error is 29.08\% when the head role is missing and 70.78\% when it is observed. Unsupported tails remain comparatively unreliable. Consequently, assigning both missing-role bits the same positive coefficient discards a consequential distinction (Table~\ref{tab:reciprocal-strata}). These are conditional outcome frequencies, not effects of deleting training support.

Relation weighting supplies a second boundary. If $w$ is the fraction of error--correct query pairs sharing a relation, exact pair accounting gives
\[
 \mathrm{AUROC}_{\rm pooled}
 =w\,\mathrm{AUROC}_{\rm within}+(1-w)\,\mathrm{AUROC}_{\rm cross}.
\]
Ties receive half credit throughout. Only 1.72\% of reciprocal FB15k-237 pairs, and 11.60\% of reciprocal CoDEx-S pairs, are within-relation. The negative orientation survives within-relation pair weighting: mean AUROCs are 0.4554 and 0.4833. It does not survive equal-relation averaging: macro values are 0.5508 and 0.5645 and exceed $1/2$ in every seed. Eligible relations depend on whether both outcomes occur. Their eligible sets and pair weights also change between predictor stages. The pooled reversal therefore does not describe a uniformly sampled relation, and changing the weighting changes the scientific question. A post-review common-set comparison holds each of four alternative relation weights fixed across stages (Appendix~\ref{sec:relation-weights}). Both pair-weight choices preserve the negative reciprocal orientation on FB15k-237 and CoDEx-S, while equal weights do not; query weights yield mixed directions on FB15k-237. Native weights are concentrated, but omitting the four largest relations also preserves the negative within-pair direction on those two graphs.

\subsection{Do counts help beyond confidence, disagreement and relation identity?}
Our \emph{Base} error head uses prediction confidence (top score, top-two margin and candidate-softmax entropy) plus log frequencies of the head, predicted tail and relation. Logistic regression (LR) and histogram gradient boosting (HGB) predict the named error outcome. \emph{Counts} adds $\log(1+n_H(h,r))$ and $\log(1+n_T(\widehat t,r))$; \emph{binary} adds their zero indicators. Fixed comparisons use $C=1$ for LR and the same 100-iteration, seven-leaf HGB budget across feature sets. Standardization and fitting use validation hash bins 0/2; reciprocal checkpoint selection uses bin 3. All score contrasts keep predictions, candidates and target sets fixed.

On pretest CoDEx-S, the four primary Counts-minus-Base AURC contrasts---LR and HGB at both training stages---all improve, in every seed. Reciprocal means are $-0.00605$ and $-0.01050$; Brier also improves. At 85\% batch coverage these correspond to 0.390/0.461 percentage-point risk reductions, or about 3.67/4.33 more correct predictions among the same 939 retained queries (Table~\ref{tab:external-codex}). The predefined directional criterion required all four mean AURC differences below $-10^{-6}$; it is descriptive, not a significance test. At risk85, however, reciprocal Counts HGB retains 2 fewer, 1 fewer, and 16 more correct predictions across the three seeds; the favorable mean does not establish a stable operating-point gain.

This external result uses 703 scorer-fitting queries, 362 threshold-fitting queries and 350 checkpoint-selection queries. Selected reciprocal epochs are 8/4/6. No CoDEx-specific predictor hyperparameter search was performed. Its published hard-negative annotations are unused, since they would define a different outcome from the setwise recovery loss. The check tests transfer of the specified recipe to one additional observation process; it does not supply a random sample of graphs.

The six original fixed ensembles average temperature-one candidate probabilities from three seeds. Their Base features replace single-model confidence with ensemble confidence; disagreement controls add the entropy gap, hard-vote disagreement and standard deviation of member probabilities for the ensemble-chosen tail. These measure algorithmic variation, not posterior uncertainty. Counts improve AURC and Brier in all twelve ensemble/head comparisons, including reciprocal ensembles with 47.30\%/46.15\% recovery on FB15k-237/WN18RR (Table~\ref{tab:ensemble-role}).

To strengthen that comparator, four LR specifications cross counts with one-hot relation identity and each select $C\in\{0.1,1,10\}$ by development-selection AURC, breaking exact ties toward smaller $C$ without refitting on selection outcomes. Standardizers and relation vocabularies use fitting queries only; unseen relations receive an all-zero indicator vector. Ensemble Base includes disagreement. This matched search tests a relation-specific intercept and limited regularization choice, not exhaustive calibration. Slopes remain shared across relations; a flexible relation-by-confidence model could absorb part of the count increment and is not evaluated here. The six CoDEx-S fits/selections preceded the first parsing of test outcomes under a supplemental plan; the other twelve tasks reuse inspected graphs. Counts improve AURC and Brier in all 18 tasks, and risk85 in 17 with one tie (Table~\ref{tab:main-relation-counts}). Improved-regime increments are considerably smaller than sampled-regime increments.

\begin{table}[t]
\centering\footnotesize
\setlength{\tabcolsep}{3pt}
\begin{tabular}{@{}llrrrr@{}}\toprule
Graph & Predictor & Runs & $\Delta$AURC & $\Delta$Brier & $\Delta$Risk85 (pp)\\\midrule
CoDEx-S & Reciprocal & 3 & -0.005117 & -0.003373 & -0.532481\\
FB15k-237 & Reciprocal & 3 & -0.004710 & -0.005834 & -0.300781\\
WN18RR & Reciprocal & 3 & -0.002238 & -0.002848 & -0.145879\\
FB15k-237 & Reciprocal ensemble & 1 & -0.001251 & -0.001472 & -0.125642\\
WN18RR & Reciprocal ensemble & 1 & -0.001144 & -0.000643 & -0.218818\\
CoDEx-S & Sampled & 3 & -0.092905 & -0.034364 & -0.461484\\
FB15k-237 & ComplEx ensemble & 1 & -0.040282 & -0.020538 & -0.331239\\
WN18RR & ComplEx ensemble & 1 & -0.002415 & -0.010559 & -0.131291\\
FB15k-237 & DistMult ensemble & 1 & -0.041300 & -0.022605 & -0.205597\\
WN18RR & DistMult ensemble & 1 & -0.004222 & -0.014467 & -0.131291\\\bottomrule
\end{tabular}
\caption{Counts added to the matching relation-ID LR head after equal-grid development selection. Negative differences favor counts. Three-run rows are descriptive seed means; a one-run row is one fixed ensemble. Complete four-specification scores and risks appear in Tables~\ref{tab:relation-identity-scores}--\ref{tab:relation-identity-risks}.}
\label{tab:main-relation-counts}
\end{table}

No representation dominates all endpoints. In the matched relation-ID comparison, risk50 worsens for both reciprocal ensembles, by 0.0388/0.0744 percentage points. In the fixed-budget reciprocal WN18RR ensemble, binary LR worsens AURC by 0.000111, binary HGB worsens Brier by 0.000222, and continuous-count HGB worsens risk85 by 0.0875 points despite improving AURC/Brier. Binary indicators add no information conditional on their counts; their observed LR effects reflect representation and fitting, while HGB scores with and without the bits are identical in the original and ensemble count controls.

Endpoint headroom also matters. Counts LR reaches the 85\%-coverage hindsight floor for all twelve original predictors but none of the six reciprocal checkpoints. Reciprocal excess risks are nevertheless only 0.411/0.321 percentage points on FB15k-237/WN18RR (Table~\ref{tab:reciprocal-score-controls}). That floor uses held-out outcomes for scale, not as an available policy; cross-predictor AURC/Brier changes additionally reflect changed error prevalence.

\subsection{Which population, target set and decision policy?}
The primary exclusion of validation-overlapping keys removes 24.86\% of eligible grouped FB15k-237 test queries. Excluded versus retained queries differ sharply in multi-answer frequency (46.89\% versus 9.80\%) and missing-head support (3.70\% versus 45.06\%; Table~\ref{tab:grouped-pool-populations}). We therefore also union held-out answers by query before assigning whole groups to partitions. This changes both the population and its recorded target sets.

The reciprocal grouped analysis discards hash bin 3 entirely, because its original validation queries selected checkpoints; all heads are newly fitted on bin 0, thresholds on bin 1, and evaluated on bin 2. Among 5,568/1,368 FB15k-237/WN18RR queries, raw-coverage AUROC is 0.3821/0.6211. The FB15k-237 reversal therefore survives this population change. Counts improve LR/HGB AURC and Brier in all six runs, but mean WN18RR LR risk85 worsens by 0.0287 percentage points. Section~\ref{sec:reciprocal-grouped} retains all heads and policy outcomes.

Within those populations, 3,381/721 queries have original-test targets. Holding their predictions, features, scores and policies fixed, unioned targets alone change 122/142/160 FB15k-237 outcomes and 8/5/8 WN18RR outcomes from incorrect to correct across seeds. This isolates label enlargement within the subset. Using original-test-only labels makes the WN18RR count-HGB AURC increment unfavorable in two seeds, and its Brier increment unfavorable in one. Population robustness and target-definition robustness are distinct claims.

Candidate restrictions change the predictor itself. Restricting original DistMult predictions to training-supported tails raises FB15k-237 accuracy from 10.56\% to 25.79\%, but lowers WN18RR from 38.16\% to 28.05\%. It excludes every recorded answer for 683/10,301 and 744/2,689 queries; all remain in the denominator. Refitted binary-role gains shrink to 0.019/0.073 percentage points at 85\% coverage (Table~\ref{tab:type-constraint}). Training support cannot serve as a universally safe hard candidate rule.

Finally, batch selection fixes $\lfloor qn\rfloor$ answers, whereas a development-fixed threshold need not transfer that answer rate. CoDEx-S thresholds accept the full boundary tie and target at least 85\% development acceptance. Reciprocal LR Base/Counts attain test rates 83.74\%/83.68\% and risks 49.06\%/48.62\%; HGB attains 85.79\%/85.13\% and 50.18\%/49.40\% (Table~\ref{tab:external-codex-threshold}). These lower risks accompany unequal answer rates. The original WN18RR randomized-boundary policy is adverse: binary augmentation increases mean accepted error from 55.18\% to 55.30\%. Neither policy provides a coverage or risk guarantee.

Saved checkpoints, fitted heads, query ledgers and numerical replays support these conditional measurements. The positive conclusion is that continuous role counts merit inclusion in specified error-ranking baselines; the negative conclusion is that raw support orientation and favorable batch metrics cannot be transferred across relations, populations or acceptance policies without evaluation.

\section{Related work}
Observed role sets and their use as approximate type constraints are established~\cite{typeconstraints}. For duplicate-free training triples, the source/target peer counts of Rossi et al.~\cite{rossipeers} equal $n_T(t,r)$/$n_H(h,r)$ for a held-out tail query. They report opposite performance associations for these directions. Shomer et al.~\cite{degreebias} also analyze tail--relation degree $n_T(t,r)$. These close precedents diagnose recorded-target link ranks; we instead measure counts at the chosen $\widehat t$ and rank that fixed prediction's observed-target error. The contribution is this operational extension and its measured regime/weighting/policy boundaries, not discovery of directional counts or their opposing associations. Reciprocal training and tensor regularization~\cite{lacroix} provide a more accurate comparison regime; deep-ensemble aggregation~\cite{deepensembles} provides an explicit disagreement baseline. The members are optimization-seed algorithms, not posterior samples.

Selective prediction orders decisions by conditional risk~\cite{franc,decisionrisk}. KG calibration distinguishes closed-world labels from truth judgments requiring additional annotation~\cite{kgcalibration}, while KGE Calibrator studies post-hoc probability calibration~\cite{kgec}. Temporal KG abstention uses a different evolving information set~\cite{selectivetkg}; predicate-conditional conformal answer sets target inclusion guarantees~\cite{condkgcp}, rather than the fraction of queries answered here. We claim no advance over these specialized calibration or guarantee methods. Additional distinctions concerning probabilistic KG models appear in Appendix~\ref{app:related}.

\section{Limitations and conclusion}
The measurements concern custom static, seen-entity query populations and incomplete recorded targets. Three optimizer seeds are not independent graph replications; continuation stages share their parent checkpoints, and the six ensembles have no independent ensemble-level repeats. Improved training changes several ingredients together, so it establishes a measured regime rather than its cause or competitive standard-benchmark performance. The external study adds one prospectively specified graph, not population-wide inference or independent execution. Grouped pooling changes targets as well as queries. No experiment establishes temporal transfer, calibrated factual truth, or a deployed risk guarantee.

The practical finding is specific: evaluate support as a learned, directional feature conditional on the actual predictor and policy. A fixed positive missing-support penalty reverses on two graphs even under pair-weighted within-relation evaluation, while equal-relation averaging gives a different conclusion. Continuous counts improve AURC and Brier beyond strengthened comparators in the recorded tasks, but the strongest-regime increments can be small and individual operating points can worsen. Saved checkpoints, ledgers and independent numerical checks make these boundaries inspectable; they do not convert descriptive effects into universal guarantees.

\clearpage
\appendix
\section{Historical results requiring reconciliation}\label{app:history}
Table~\ref{tab:legacy} preserves a compact record of the earlier manuscript's reported results. They are not newly validated measurements and are not results for the role-specific extension.
\begin{table}[ht]
\centering\small
\begin{tabular}{lrr}\toprule
Reported support-diagnostic AUROC & FB15k-237 & YAGO3-10\\\midrule
Entity score (labeled frequency in original) & 0.542 & 0.824\\
Coverage & 0.935 & 0.760\\
Fixed average & 0.951 & 0.892\\
Learned CAGP mixture & 0.986 & 0.942\\\bottomrule
\end{tabular}
\caption{Values transcribed from Table 1 of arXiv:2512.22318v2; reproduction pending. The label rule itself is a necessary comparator.}\label{tab:legacy}
\end{table}
\begin{samepage}
The original reports two incompatible FB15k-237 class-count triples: $(2223,5193,13050)$ in Table 2/Appendix C and $(1317,5678,13471)$ in Appendix A.4.1. Both sum to 20,466; they cannot describe the same fixed partition. Appendix B refers to an ICEWS14 result absent from the cited main table.

\begin{proposition}[A consistency bound for selective accuracy]
For fixed predictions on a fixed evaluation set, full-set accuracy $a$ and retained fraction $q>0$ imply retained accuracy $a_q\leq\min(1,a/q)$.
\end{proposition}
\begin{proof}
There are $an$ correct predictions in total. A retained set of $qn$ items contains at most $an$ correct predictions, so $a_q qn\leq an$.
\end{proof}
\end{samepage}
The reported pair $a=0.623$, $q=0.85$ permits at most $0.73294$ retained accuracy, below the reported $0.784$. If models or evaluation populations differ, this comparison does not isolate abstention. The 43\% error-reduction headline should be withheld until the prediction ledger explains the discrepancy.

\section{Detailed information contract and metric conventions}\label{app:contract}
\subsection{Outcome, candidates, and permitted information}
For a tail query $(h,r,?)$, a frozen predictor selects $\widehat t$ from a specified candidate set using a fixed prediction tie rule. Let $T_{h,r}$ contain the recorded held-out answers. The outcome is the observed-target error
\[
 L(h,r)=\ind\{\widehat t\notin T_{h,r}\}.
\]
Queries, not individual answer triples, receive equal weight. This is a recovery task: an unrecorded true answer can be counted as an error, so $L$ is not a ground-truth falsehood label. A paired score comparison keeps the query keys, target sets, predictions, and candidates identical. Changing the predictor or candidate policy defines a separate comparison.

Every feature used to decide whether to answer must be available before revealing $T_{h,r}$. Our candidates and support counts use training data only; endpoint features are evaluated at $(h,r,\widehat t)$. Scoring the hidden true tail instead can leak the answer even when the counts themselves are training-only. Appendix~\ref{app:diagnostics} gives an exact two-query example in which this changes AUROC from $1/2$ to $1$. It demonstrates possibility, not a defect established in another method or the historical experiments. Held-out-fact filtering likewise cannot determine an operational prediction or its features. For triple classification both endpoints are inputs, so the information contract differs.

\subsection{Training counts and role-support representations}
Let $\mathcal T_{\rm tr}$ be the training triples. Define the entity count
\begin{equation}
 f(e)=\sum_{(h,r,t)\in\mathcal T_{\rm tr}}\ind\{e=h\ \text{or}\ e=t\}.
\end{equation}
A self-loop contributes once to $f(e)$. For directed relations, define separate head and tail evidence:
\begin{align}
 n_H(e,r)&=\sum_{(h,r',t)\in\mathcal T_{\rm tr}}\ind\{h=e,r'=r\},\\
 n_T(e,r)&=\sum_{(h,r',t)\in\mathcal T_{\rm tr}}\ind\{t=e,r'=r\},\qquad
 c_a(e,r)=\ind\{n_a(e,r)>0\},\ a\in\{H,T\}.
\end{align}
Separating roles extends the original manuscript; the underlying observed domain/range sets are established type-support constructions \cite{typeconstraints}. The original manuscript pooled head and tail roles; its numerical results cannot be assigned to this extension without a rerun.

The normalized coverage score is
\begin{equation}\label{eq:coverage}
 U_C(h,r,t)=1-\frac{c_H(h,r)+c_T(t,r)}{2}\in\{0,\tfrac12,1\}.
\end{equation}
Missing coverage is evidence of an unobserved training configuration, not evidence that the triple is false. Under an open-world interpretation, absent facts may be true.

For the predicted triple, write $b_H=1-c_H(h,r)$ and $b_T=1-c_T(\widehat t,r)$. The scalar score $U_C=(b_H+b_T)/2$ discards their direction; our logistic and nonlinear comparisons also use them separately. The identities $b_a=\ind\{n_a=0\}$ show that adding these bits to the exact role counts adds no information. A gain for a fitted logistic model can nevertheless reflect its feature representation, regularization, and estimation. Frequency is a count baseline, not a measurement of learned embedding variance; the original Gaussian-score motivation is retained only in Appendix~\ref{app:theory}.

\subsection{Selection, probability scoring, and ties}
At requested batch coverage $q$, retain $k=\lfloor qn\rfloor$ of the $n$ predictions with the lowest error score. The selective risk is their observed-target error fraction. This is a batch-ranking endpoint; an online threshold fixed on development data instead has a test answer rate that can change. We evaluate the latter policy separately and report its achieved rate.

Coverage scores take only three values, so ties must be handled explicitly. Sorting tied scores by correctness would artificially improve selective risk. For a tied block of $m$ queries with $e$ errors, suppose $b$ lower-score queries with $E_b$ errors have already been retained. Uniformly selecting $j$ queries from the block gives
\begin{equation}
 \E[R((b+j)/n)]=\frac{E_b+je/m}{b+j},\qquad 1\leq j\leq m.
\end{equation}
The denominator is fixed, and each error in the block is selected with probability $j/m$. Our evaluator averages this random tie rule analytically, avoiding dependence on file order. It reports actual retained counts $k=\lfloor qn\rfloor$ and the discrete summary $n^{-1}\sum_{k=1}^n\E[R(k/n)]$. This stated convention avoids ambiguity about interpolation at zero coverage; it is not a claim of an unbiased population AURC estimator.

Each input row must identify a unique query, its fixed-predictor correctness, and its uncertainty score. The output records the input hash, denominator, prevalence, tie-aware error AUROC, full accuracy, and selective risks. An optional estimated error probability supports a separately named Brier score. A single-class file has undefined AUROC and is reported as such. This contract checks arithmetic and population consistency; it cannot certify how predictions, candidates, or calibration were produced.

A probability forecast $s_i\in[0,1]$ for the named error outcome is assessed by $n^{-1}\sum_i(L_i-s_i)^2$, the Brier score. It need not already be calibrated for this score to be meaningful. A raw novelty score has no automatic probability interpretation. AUROC, threshold-grouped AP, and the paired-inference conventions are specified in Appendix~\ref{app:metrics}.

\subsection{What the supporting results do and do not establish}
Table~\ref{tab:theory-map} separates information-level statements from the implemented estimators. The proofs in Appendix~\ref{app:theory} adapt standard probability and conditional-risk arguments \cite{franc,decisionrisk}. None gives an operational performance guarantee for our fitted logistic or HGB scorer: its approximation error, finite-data estimation, and distribution shift are not bounded by these experiments. In particular, the primary baseline already contains relation-dependent confidence and relation counts, so its measured improvement is not an instantiation of an entity-only indistinguishability theorem.

\begin{table}[ht]
\centering\small
\begin{tabularx}{\linewidth}{@{}p{0.25\linewidth}XX@{}}\toprule
Supporting statement & Premise and outcome & Role in the executed study\\\midrule
Fixed-feature TV bound (Proposition~\ref{prop:tv}) & Fixed endpoint-feature distributions; tie-aware novelty AUROC. & Scope diagnostic; not a bound on predicted-endpoint LR/HGB error ranking.\\
Brier projection (Eq.~\eqref{eq:brier-context}) & Exact conditional error probabilities under nested information. & Defines oracle information value; does not assert fitted-score improvement.\\
Optimal selection and plug-in regret & Equal population coverage; exact or controlled error-probability approximation. & Explains oracle ordering; does not certify development-threshold transfer.\\
Mixture separation (Proposition~\ref{prop:mix}) & Complete separation and bounded adverse score differences. & Idealized sufficient condition; not assumed for fitted directional scorers.\\
Count-to-bit determinism & Exact role counts and their zero indicators. & Directly motivates the Counts LR and count-HGB controls.\\\bottomrule
\end{tabularx}
\caption{A map from supporting arguments to the actual estimator comparisons. The premises are not supplied merely by observing favorable benchmark metrics.}\label{tab:theory-map}
\end{table}

\section{Additional related-work distinctions}\label{app:related}
UKGE learns from confidence-valued facts \cite{ukge}; its uncertainty target differs from posterior parameter uncertainty. BEUrRE applies relation-specific affine transformations to entity boxes \cite{beurre}, so it should not be characterized as universally relation-agnostic. The paper cited in the original as a Chen et al. Gaussian-process method is actually PERM by Choudhary et al. \cite{perm}. A Gaussian representation does not establish a Gaussian-process prior. These distinctions are necessary for fair baselines and accurate novelty claims.

Selective prediction and calibration in knowledge graphs already have dedicated methods. Safavi et al.\ \cite{kgcalibration} distinguish closed-world calibration labels from open-world labels requiring additional ground truth. Our observed-target loss is explicitly a recorded-fact recovery outcome and does not resolve that open-world annotation problem. Hou et al.\ \cite{selectivetkg} study abstention in temporal KG reasoning using current prediction certainty and historical correctness; our static support-count study addresses a different information set and does not replace their temporal evaluation. KGE Calibrator \cite{kgec} studies post-hoc probability calibration while preserving link-ranking behavior. Our logistic error heads are simple baselines for incremental feature assessment, not a claimed advance over specialized calibration methods.

Predicate-conditional conformalized answer sets \cite{condkgcp} target inclusion of the true answer in a returned set, including predicate-conditioned guarantees under their assumptions. That notion of coverage differs from both training-role support and the fraction of queries answered in our selective-risk curves. We provide no conformal coverage guarantee. The present contribution is a reproducible assessment of when established role-support features help a specified error-ranking model, together with explicit failures of extrapolation across candidate and acceptance policies.

The predictor-quality extension adapts reciprocal training and tensor regularization from Lacroix et al.\ \cite{lacroix}, with a low-rank warm start and a distinct fixed budget; it does not reproduce their published configuration. Our ensemble diagnostic uses the equally weighted probability-mixture convention of deep ensembles \cite{deepensembles}. The entropy difference also appears in Bayesian acquisition functions \cite{galactive}, but three optimization seeds are algorithmic members, not posterior samples. Its use here supplies a concrete disagreement comparator, not an epistemic/aleatoric decomposition.

The supporting fixed-feature bound concerns the information retained by a chosen score. It does not rule out relation-aware ensembles, calibrated triple probabilities, or model architectures that use relational context. The proposed contribution is an auditable relation-evidence diagnostic and evaluation framework, not a claim that probabilistic KG methods as a class cannot represent relational uncertainty.

\section{Complete empirical methods and results}\label{sec:new-experiments}
We study FB15k-237 \cite{fb237} and WN18RR \cite{wn18rr} through two base architectures, a higher-accuracy training regime, and fixed ensembles. The original study trains six DistMult \cite{distmult} models; six parameter-matched ComplEx models provide an architecture check. A later development-only pilot precedes six additional warm-start reciprocal ComplEx stages. These are 12 from-scratch predictors plus 6 continued training stages, not 18 independent architecture trials. Finally, one three-member ensemble is fixed for each of the three model/stage configurations on each dataset. These are new real-data measurements, not reproductions of the original Gaussian embedding, CAGP implementation, or historical results. The original protocol and primary comparison were recorded before its test scoring. Subsequent analyses were fixed before their immediate executions but after earlier results; none is an independent preregistration or confirmation.

\subsection{Prediction task, population, and fitting}
The task is static missing-fact tail prediction. The vocabulary contains only entities and relations observed in training. We exclude held-out triples with unseen identifiers and remove targets already observed for the same query in training. Within each held-out split, all remaining triples sharing $(h,r)$ form one query with a set $T_{h,r}$ of observed targets. Test query keys also appearing in validation are excluded. A prediction is correct when $\widehat t\in T_{h,r}$; hence the loss is $\ell(h,r)=\ind\{\widehat t\notin T_{h,r}\}$. Each query has equal weight regardless of its number of recorded targets. This grouping avoids treating different recorded answers to one query as contradictory single-answer labels.

At decision time, all training entities are candidates except tails already known for that query from training. Neither validation nor test facts filter the candidates. All endpoint features use the predicted tail, and a score tie is resolved by the first entity in the sorted training vocabulary. The resulting test populations contain 10,301 FB15k-237 queries and 2,689 WN18RR queries. These are selected subsets of the published test splits, rather than standard filtered link-prediction populations. In particular, an unrecorded true answer is counted as a failure to recover an observed target, not established falsehood. No temporal ordering is inferred from these static splits. Table~\ref{tab:attrition} records the population changes. Removing validation-overlapping query keys excludes 24.86\% of eligible grouped FB15k-237 test queries and 4.44\% on WN18RR; these subsets can differ in difficulty and answer multiplicity from the full benchmarks.

\begin{table}[t]
\centering\small
\begin{tabular}{lrr}
\toprule
Population stage & FB15k-237 & WN18RR\\
\midrule
Published test triples & 20,466 & 3,134\\
Triples excluded: unseen training IDs & 28 & 210\\
Targets excluded: already in training & 0 & 0\\
Grouped test queries before overlap removal & 13,709 & 2,814\\
Queries excluded: overlap with validation & 3,408 & 125\\
Evaluated test queries & 10,301 & 2,689\\
Observed targets in evaluated queries & 11,579 & 2,746\\
\bottomrule
\end{tabular}
\caption{Test-population accounting, identical across seeds and models. Rows explicitly distinguish triples/targets from grouped queries; they cannot be subtracted across units. Query-overlap removal also removes all associated observed targets. Validation grouping precedes its disjoint fitting/selection partition.}
\label{tab:attrition}
\end{table}

For each dataset we use seeds $17,29,43$, 64-dimensional embeddings, 30 fixed epochs, batch size 1,024, 16 negative corruptions per positive, and Adam with learning rate $0.003$. For positive-triple scores $s^+$ and sampled-negative scores $s^-$, the loss averages $\operatorname{softplus}(-s^+)$ and $\operatorname{softplus}(s^-)$ separately and adds $10^{-4}$ sampled-embedding mean-square regularization. Head or tail corruption proposals are uniform before rejecting known training positives; rejection can alter the accepted head/tail proportions. Unobserved true triples may still be sampled as negatives. We use the final checkpoint without selecting an epoch from test or development performance. This fixed CPU budget does not establish convergence or competitive embedding accuracy.

Validation query keys are divided by SHA256 parity of the original query string into disjoint fitting and selection partitions, using only query identifiers, without consulting outcomes. Their sizes are 6,027/6,045 for FB15k-237 and 1,370/1,338 for WN18RR. Standardization, logistic scorers, and the rarity-score empirical CDF use only the fitting partition. The mixture weight is selected on the selection partition using risk at 85\% batch coverage over $\alpha\in\{0,0.1,\ldots,1\}$, taking the smallest weight on an exact tie. Every test run uses the frozen predictor and scorers.

\subsection{Comparisons and recorded endpoint}
Nine scores are compared on each fixed set of predictions. The unsupervised scores are negative top-two score margin, normalized endpoint rarity, role coverage, their fixed equal-weight mixture, and their development-selected mixture. Rarity is a decreasing function of the minimum training frequency of the known and predicted endpoints, mapped through the fitting-partition empirical CDF; it is not learned posterior variance.

The four logistic error predictors form nested feature sets. Confidence logistic uses the top score, top-two margin, and softmax entropy over the permitted candidates. Base LR additionally includes both endpoint log-counts \emph{and the relation log-count}. This confidence-and-count baseline is labeled \emph{entity-count logistic} in the archived original table (Table~\ref{tab:from-scratch-results}); it is not an entity-only score. Base + binary LR adds the two indicators $1-c_H(h,r)$ and $1-c_T(\widehat t,r)$ with separately fitted coefficients of unrestricted sign. It does not use their scalar equal-weight average $U_C$ or the convex mixture in Eq.~\eqref{eq:mix}. Counts + binary LR further adds $\log(1+n_H(h,r))$ and $\log(1+n_T(\widehat t,r))$. Each model uses the same fitting queries, standardization fitted there, and $L_2$-regularized logistic regression with $C=1$ and no class reweighting.

The recorded primary contrast is Base + binary LR minus Base LR in observed-target error risk at 85\% batch coverage. The predictor and evaluation population are identical within each comparison. Tied score blocks use the analytic random-order expectation described above. We report seed means and ranges, with per-seed values, secondary metrics, fitted coefficients, checkpoints, and prediction ledgers archived in the repository. The three seeds share a graph and test queries; their range is not a confidence interval. We do not conduct significance tests or treat seeds as independent replications of a graph population.

The complete original nine-score comparison is retained in Table~\ref{tab:from-scratch-results} in Appendix~\ref{app:empirical-detail}.

\subsection{A matched-budget asymmetric predictor check}
After the DistMult results, we recorded a further model check using ComplEx \cite{complex}, whose score is $\operatorname{Re}\sum_j h_j r_j\overline{t_j}$. Its 32 complex components use 64 real coordinates per embedding, matching DistMult's parameter count exactly (943,488 parameters on FB15k-237 and 2,596,480 on WN18RR). We retained all three seeds, 30 fixed epochs, training objective, original candidate rule, query populations, development partitions and nine scorers. The implementation and configuration were recorded before these six additional runs. No architecture or epoch search was performed. This is post-primary-result sensitivity analysis, not independent confirmation; score geometry and optimization change along with the ability to represent asymmetry.

Table~\ref{tab:complex-results} in Appendix~\ref{app:empirical-detail} reports the matched-budget architecture results.

Coverage augmentation reduces risk relative to the confidence-and-count logistic model in all six ComplEx runs, by a mean 0.47 percentage points on FB15k-237 and 0.35 points on WN18RR. Thus the positive increment also occurs in an asymmetric-capable architecture in this budget; it cannot be attributed exclusively to a mandatory symmetric score. This is not a claim that the learned model fully exploits asymmetry. Mean full-coverage accuracy is 10.05\% and 37.47\%, respectively, 0.51 and 0.68 points lower than the corresponding DistMult means. This architecture check does not itself supply a stronger predictor; Section~\ref{sec:reciprocal} evaluates a separately developed higher-accuracy regime. Saved checkpoints and development/test ledgers support the within-predictor comparisons.

\subsection{Measured gains and unfavorable comparisons}
Base + binary LR reduces the primary risk relative to the confidence-and-count baseline in all six DistMult runs. The mean paired reductions are 0.54 percentage points on FB15k-237 (seed range 0.45--0.69) and 0.28 points on WN18RR (0.18--0.44). Absolute retained-query risks remain 87.64\% and 55.92\%, respectively, while full-coverage observed-target accuracy averages 10.56\% and 38.16\%. High retained risk reflects the base predictor's errors and does not, by itself, imply poor error ranking. With $c$ correct predictions and $k=\lfloor0.85n\rfloor$ retained queries, the hindsight minimum risk is $\max(0,1-c/k)$. For DistMult on FB15k-237 its seed mean is 87.577\%, leaving the 88.174\% baseline only 0.598 percentage points of improvement possible. Coverage augmentation closes 89.8\% of this mean headroom and retains a mean 99.51\% of all correct predictions. This oracle context limits what the 85\% endpoint can distinguish; it is not a deployable selection rule.

Improvement over the chosen nested baseline does not establish superiority over every simple score. On WN18RR, the score-margin baseline has lower mean risk, 55.74\%, than Base + binary LR's 55.92\%. Counts + binary LR, an exploratory extension, has mean risks 87.58\% and 55.10\%. This comparison motivates retaining counts as a substantive baseline rather than attributing all useful contextual information to binary coverage.

The mixture results also constrain the interpretation. All three FB15k-237 runs select $\alpha=0$, yielding coverage alone; the fixed mixture has higher mean risk than coverage alone (88.20\% versus 87.93\%). Thus these runs do not support empirical complementarity of rarity and coverage at the primary endpoint, despite the sufficient-condition result for a constructed population. WN18RR selects $\alpha=0.8$ in all three runs, but its selected mixture still has higher mean risk than the margin baseline (56.74\% versus 55.74\%). A useful weight is dataset- and endpoint-dependent.

\subsection{Role representation and the strongest scorer comparisons}
A further analysis fixed before its scorer fits examines ten error models on all twelve frozen predictors and their original query populations. It adds head-only and tail-only indicators, continuous role counts without binary indicators, relation-identity one-hot features, and histogram gradient boosting (HGB). Only the original development-fit queries train these scorers; no hyperparameter search is performed. Logistic settings remain unchanged. HGB uses 100 iterations, at most seven leaves, learning rate $0.1$, $L_2$ penalty one, and no early stopping. This post-primary-result analysis tests dependence on the comparator and function class, not independent replication.

In Table~\ref{tab:score-ablation-results}, \emph{Counts LR} adds both continuous log-role counts to the existing confidence, entity-count and relation-count features; it is not a counts-only link predictor. \emph{Counts + binary LR} additionally includes the missing-role indicators and corresponds to the earlier role-count-logistic row in the archived table. This distinction isolates the incremental value of explicit zero-count features. Because a missing-role indicator is a deterministic function of its role count, it supplies no additional information conditional on that count. It can nevertheless change the approximation available to a penalized linear logistic model.

\begin{table}[t]
\centering
\small
\begin{tabular}{lrrrr}
\toprule
 & \multicolumn{2}{c}{DistMult} & \multicolumn{2}{c}{ComplEx} \\
Scorer & FB15k-237 & WN18RR & FB15k-237 & WN18RR \\
\midrule
\multicolumn{5}{l}{Risk at 85\% coverage (percent)} \\
Base LR & 88.174 & 56.193 & 88.704 & 56.791 \\
Base + binary LR & 87.638 & 55.915 & 88.231 & 56.441 \\
Counts LR & 87.577 & 55.098 & 88.178 & 55.901 \\
Counts + binary LR & 87.584 & 55.098 & 88.178 & 55.901 \\
Relation ID LR & 87.797 & 55.478 & 88.403 & 56.411 \\
Relation ID + binary LR & 87.603 & 55.332 & 88.190 & 56.178 \\
Counts HGB & 87.580 & 55.098 & 88.178 & 55.915 \\
Hindsight oracle & 87.577 & 55.098 & 88.178 & 55.901 \\
\midrule
\multicolumn{5}{l}{Discrete AURC / Brier score (unscaled)} \\
Base LR & 0.774 / 0.080 & 0.340 / 0.121 & 0.783 / 0.077 & 0.275 / 0.051 \\
Base + binary LR & 0.726 / 0.064 & 0.309 / 0.108 & 0.733 / 0.061 & 0.271 / 0.044 \\
Counts LR & 0.699 / 0.051 & 0.310 / 0.102 & 0.708 / 0.049 & 0.266 / 0.038 \\
Counts + binary LR & 0.696 / 0.049 & 0.294 / 0.092 & 0.707 / 0.047 & 0.265 / 0.034 \\
Relation ID LR & 0.741 / 0.067 & 0.261 / 0.036 & 0.753 / 0.066 & 0.267 / 0.031 \\
Relation ID + binary LR & 0.700 / 0.052 & 0.259 / 0.034 & 0.715 / 0.052 & 0.264 / 0.027 \\
Counts HGB & 0.687 / 0.043 & 0.253 / 0.015 & 0.700 / 0.043 & 0.259 / 0.013 \\
\bottomrule
\end{tabular}
\caption{Exploratory frozen-predictor scorer ablations: means over three seeds; lower is better. Base includes confidence and entity/relation counts. Counts adds both continuous log-role counts; Relation ID adds training-relation one-hot features; binary adds both missing-role indicators. HGB is histogram gradient boosting on the Counts features. HGB with binary indicators gives identical scores and is omitted as a duplicate row. The hindsight oracle uses test outcomes solely to show the feasible risk floor. All ten methods, seeds and ranges are archived; no confidence intervals are implied.}
\label{tab:score-ablation-results}
\end{table}

Counts LR attains the hindsight minimum risk at 85\% coverage in all twelve original runs, retaining every correct prediction. Adding binary indicators ties this risk in eleven runs and worsens it in one DistMult/FB15k-237 run. The endpoint is therefore saturated for this stronger comparator; it cannot support a binary-coverage advantage beyond continuous role counts. This does not mean the indicators never help: they improve logistic AURC and Brier score in all twelve runs. At 50\% coverage they instead worsen risk in all six FB15k-237 runs, while improving five WN18RR runs and tying one. Score quality depends on the loss and coverage being assessed.

HGB on continuous role counts, with and without the binary indicators, produces identical fitted, development-selection and test scores in every run. For this fitted tree model the deterministic features are redundant, although that is not a guarantee for every algorithm or configuration. An unbinned axis-aligned tree can express a zero-count indicator split by splitting the existing log-count at zero and, if needed, exchanging the child branches. Adding such an indicator therefore does not enlarge that tree function class at a fixed depth. Histogram binning, regularization and fitting choices still affect the learned model, so exact fitted-score equality remains an empirical observation here. Adding the indicators to a logistic model with relation identity still reduces mean 85\%-coverage risk by 0.19/0.15 percentage points for DistMult and 0.21/0.23 for ComplEx on FB15k-237/WN18RR. The two indicators are not uniformly complementary: on DistMult/WN18RR, the tail-only logistic model has lower mean risk than the two-indicator model (55.80\% versus 55.92\%). Together these results support comparator-dependent value of an established representation of support, rather than a distinct uncertainty component or a universal benefit from combining directions.

\paragraph{Holding the nonlinear family and hyperparameter budget fixed.}
A subsequent analysis fits HGB on the base features alone and on base plus binary roles, using the same fitting queries, 100-iteration/seven-leaf budget and other settings as the existing role-count HGB. This separates adding role features from switching scorer families. Adding features changes the split-search space, so effective function-class capacity is not exactly held constant. The configuration was fixed before these fits, but after the preceding results, and the comparison remains exploratory.

\begin{table}[t]
\centering\small
\begin{tabular}{lllrrr}\toprule
Predictor & Dataset & Error-head features & Risk85 (\%) & AURC & Brier\\\midrule
DistMult & FB15k-237 & Base & 87.645 & 0.73356 & 0.06561\\
DistMult & FB15k-237 & Base + binary roles & 87.588 & 0.70008 & 0.05215\\
DistMult & FB15k-237 & Base + role counts & 87.580 & 0.68720 & 0.04333\\
\addlinespace
DistMult & WN18RR & Base & 55.230 & 0.25545 & 0.02219\\
DistMult & WN18RR & Base + binary roles & 55.098 & 0.25319 & 0.01733\\
DistMult & WN18RR & Base + role counts & 55.098 & 0.25268 & 0.01474\\
\addlinespace
ComplEx & FB15k-237 & Base & 88.239 & 0.74735 & 0.06470\\
ComplEx & FB15k-237 & Base + binary roles & 88.178 & 0.71275 & 0.05141\\
ComplEx & FB15k-237 & Base + role counts & 88.178 & 0.69952 & 0.04268\\
\addlinespace
ComplEx & WN18RR & Base & 55.945 & 0.26159 & 0.02194\\
ComplEx & WN18RR & Base + binary roles & 55.915 & 0.25932 & 0.01673\\
ComplEx & WN18RR & Base + role counts & 55.915 & 0.25889 & 0.01344\\
\addlinespace
\bottomrule\end{tabular}
\caption{Exploratory error-head controls using the same HGB family and hyperparameter budget (100 iterations, at most seven leaves per tree) and development-fit partition. Adding features changes the split-search space, so effective capacity is not identical. Base features are prediction confidence and entity/relation frequencies; role-count rows add two log role counts. Values are means over three seeds, not confidence intervals. Risk85 is reported in percent; AURC and Brier are unscaled.}
\label{tab:nonlinear-controls}
\end{table}

The base HGB already retains a mean 99.45--99.90\% of all correct predictions at 85\% coverage across the four model/dataset groups. Binary augmentation reduces mean risk by only 0.029--0.131 percentage points at this nearly saturated endpoint. Both binary-role and role-count additions nevertheless reduce mean AURC and Brier score in every group compared with the base HGB with the same family and hyperparameter budget (Table~\ref{tab:nonlinear-controls}). Thus the role-feature benefit on these metrics is not solely a consequence of comparing a flexible scorer against a linear baseline. Continuous counts are not the best representation at every coverage: on ComplEx/FB15k-237 at 50\% coverage, binary-role HGB has mean risk 79.929\%, slightly below role-count HGB's 79.942\%. These descriptive differences support task- and metric-specific value, not a uniformly superior feature set or a significance claim.

\begin{figure}[t]
\centering
\includegraphics[width=\linewidth]{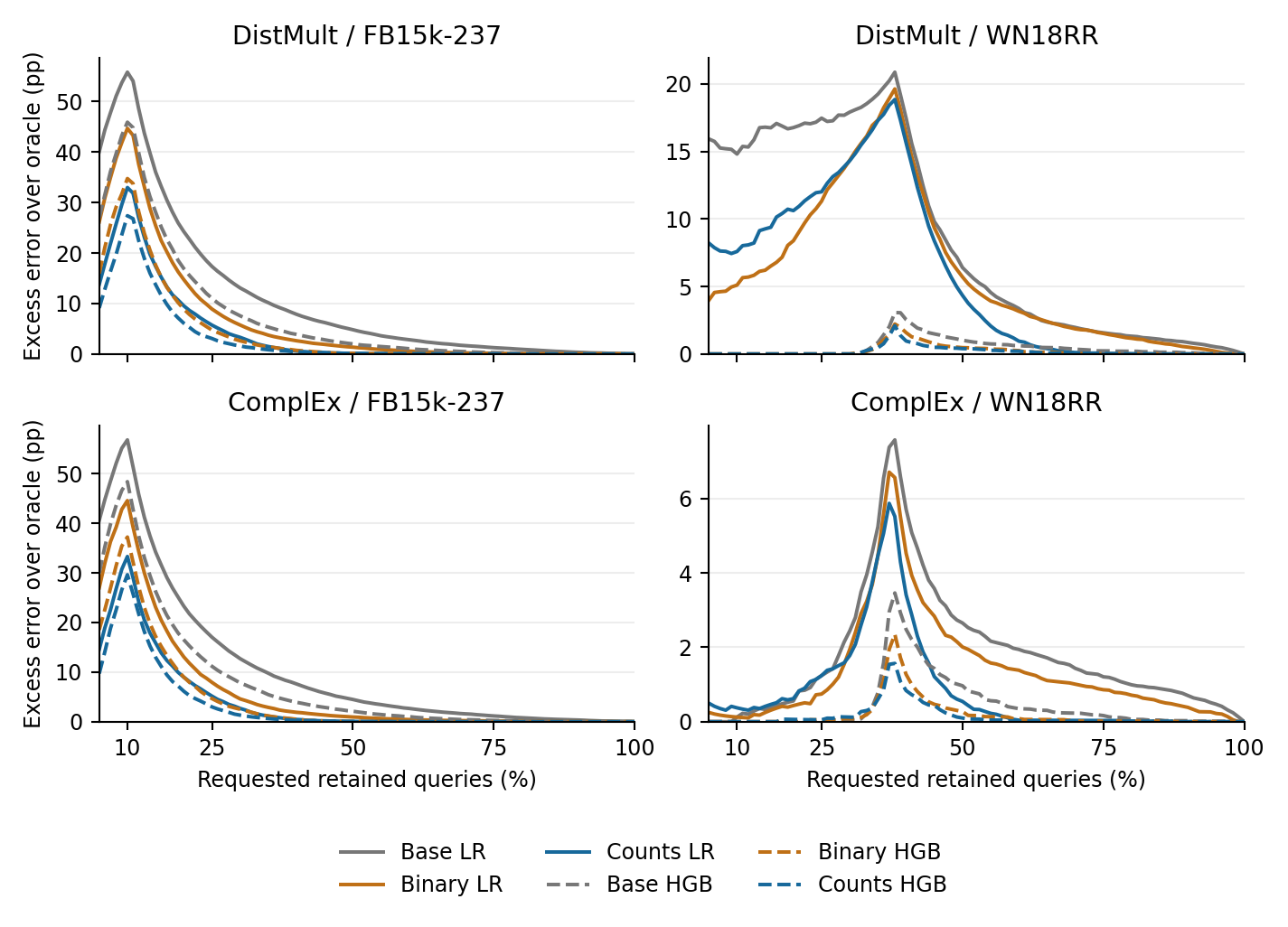}
\caption{Risk above the same-prediction hindsight floor, against batch coverage, for six LR/HGB comparisons. Lines are equal-seed means over three training runs; no sampling interval is implied. Zero is the minimum for those fixed predictions, not a deployable score. Panel scales differ. The original primary-score risk curves are retained in Appendix~\ref{app:empirical-detail}.}\label{fig:scorer-excess-risk}
\end{figure}

\paragraph{Ranking headroom across the curve.}
With $c$ correct predictions among $n$ fixed queries, the hindsight floor at rank $k$ is $R_{\rm or}(k/n)=\max(0,1-c/k)$. Its discrete AURC is $n^{-1}\sum_{k=1}^n R_{\rm or}(k/n)$. Subtracting this floor from a scorer's curve measures remaining finite-ledger ranking headroom; it is not Bayes regret or a deployable policy. Figure~\ref{fig:scorer-excess-risk} displays the strongest relevant comparators without choosing a replacement primary endpoint.

For DistMult/FB15k-237, DistMult/WN18RR, ComplEx/FB15k-237, and ComplEx/WN18RR respectively, mean oracle AURCs are $0.657169$, $0.250933$, $0.668720$, and $0.257574$. Base-HGB excess AURCs are $0.076391$, $0.004513$, $0.078629$, and $0.004015$; adding continuous role counts reduces them to $0.030035$, $0.001745$, $0.030802$, and $0.001317$. Role counts improve the measured curves, but WN18RR starts with far less absolute ranking headroom. Raw metrics and the recorded 85\% contrast remain the reported outcomes; these oracle comparisons explain their scale.

\subsection{A more accurate predictor changes the support--error relationship}\label{sec:reciprocal}
The matched-budget ComplEx check does not improve prediction. We therefore conducted a bounded development-only pilot on FB15k-237 seed 17, evaluating additional epochs zero and two without reading or scoring test queries, then froze the full training recipe. The pilot checkpoint was not reused. Each of the six original ComplEx checkpoints separately initializes an eight-epoch continued training stage with reciprocal triples, all-entity softmax cross entropy, tensor $N3$ regularization, and a fresh Adagrad optimizer. This adapts the recipe of Lacroix et al.\ \cite{lacroix}, not their published high-rank experiment.

Validation groups are assigned by original-identifier SHA256 modulo four: bins 0/2 fit error scorers, bin 1 selects scalar-mixture weights, and bin 3 selects the predictor checkpoint. All eight additional epochs run; epochs $0,2,4,6,8$ are evaluated on bin 3, and the earliest epoch with maximum observed-target accuracy is selected. Scorer-fit outcomes do not choose that epoch. This separation governs the new analysis, but the underlying validation and test graphs had already been inspected earlier. The original test queries, observed target sets and train-known candidate masks are unchanged. Counts always use the original training triples, never their reciprocal augmentation.

The 32 complex components are retained. Each inverse relation starts as its parent's complex conjugate and is then optimized independently, adding 64 real parameters per original relation. Batch size is 512 and Adagrad learning rate is $0.1$. The $N3$ term is $0.01/B$ times the sum of cubed complex-coordinate magnitudes over the $B$ sampled head, relation and tail factors; unlike the $\lambda/3$ convention in \cite{lacroix}, there is no factor $1/3$ in our coefficient. Single-target softmax makes other observed or unobserved true tails compete as labels. These choices change the objective, optimizer, auxiliary parameters and training budget together, so the result cannot isolate a reciprocal-training mechanism.

\begin{table}[t]
\centering
\small
\begin{tabular}{lrrrr}
\toprule
Dataset & Epochs & Accuracy & Change (pp) & Tail-missing (\%) \\
\midrule
FB15k-237 & 4/6/6 & 44.35 [44.22, 44.45] & 34.31 [33.94, 34.68] & 8.54 [8.24, 8.83] \\
WN18RR & 4/4/8 & 44.95 [44.66, 45.30] & 7.47 [7.03, 7.92] & 23.37 [22.02, 26.03] \\
\bottomrule
\end{tabular}
\caption{Predictor-quality sensitivity after original results. Selected epochs correspond to seeds 17/29/43 and use only the checkpoint-selection development partition. Accuracy and predicted-tail missing-role fractions are percentages; changes are against each original ComplEx checkpoint on identical test queries. Entries show mean [minimum, maximum], not confidence intervals. Multiple training choices and inverse-relation parameters change together.}
\label{tab:reciprocal-predictor}
\end{table}

The selected epochs are 4/6/6 on FB15k-237 and 4/4/8 on WN18RR (seeds 17/29/43). Checkpoint-selection accuracy rises from 9.50--10.86\% to 37.94--38.00\% on FB15k-237 and from 32.06--34.20\% to 38.96--39.42\% on WN18RR. Mean test observed-target accuracy then rises from 10.05\% to 44.35\% and from 37.47\% to 44.95\%, respectively (Table~\ref{tab:reciprocal-predictor}). This establishes a higher-accuracy regime for the defined task, without establishing convergence or a standard filtered-benchmark gain.

\paragraph{A sign reversal for fixed positive coverage scoring.}
On FB15k-237, the scalar missing-role score $U_C$ has mean error AUROC $0.3639$, below $0.5$ in every continued-training run, compared with $0.6893$ for the original ComplEx checkpoints. Its mean AURC is $0.6563$, worse than the margin score's $0.2890$. The comparison is descriptive across different fixed predictions, not an isolated causal effect of improving accuracy. It nevertheless refutes an invariant interpretation in which higher missing-role coverage necessarily ranks errors higher across these regimes.

The joint strata describe that failure; the following ranges span the three seeds. Mean missing-tail frequency falls from 55.50\% to 8.54\% on FB15k-237. Supported-tail predictions have 52.15--52.64\% error, versus 90.66--91.15\% for unsupported tails. Within the much larger supported-tail population, however, head-missing queries have only 28.36--29.63\% error, compared with 70.30--71.15\% when the head role is observed. A fixed positive penalty on both directions loses this distinction. On WN18RR the marginal tail association also differs: unsupported-tail predictions have 45.44--54.43\% error, below the supported-tail rates of 55.35--57.97\% in every seed. Their mean frequency falls from 50.01\% to 23.37\%. These are conditional rates for changing prediction populations; they do not identify effects of removing support or establish the truth of unrecorded answers. Table~\ref{tab:reciprocal-strata} in Appendix~\ref{app:empirical-detail} supplies all denominators. Original-predictor strata and relation-composition sensitivity appear there as well.

\paragraph{Which relation comparisons reverse?}
The pooled AUROC compares an error query with a correct query, usually from different relations. Let $w$ be the fraction of all error--correct pairs sharing a relation. Partitioning those pairs gives the exact finite-ledger identity
\[
 \mathrm{AUROC}_{\rm pooled}
 =w\,\mathrm{AUROC}_{\rm same}+(1-w)\,\mathrm{AUROC}_{\rm cross}.
\]
Each component uses the same half-credit tie convention; a component with no pairs is undefined and contributes zero mass. Table~\ref{tab:relation-pair-diagnostic} reports this post-result decomposition alongside an equal-weight mean over relations containing both outcome classes. On reciprocal FB15k-237, only 1.72\% of the error--correct pairs share a relation. Mean cross-relation AUROC is 0.3623, while pair-weighted within-relation AUROC is 0.4554; the latter is below $1/2$ in each seed. Thus the reversal is not solely an artifact of cross-relation comparison. However, the equal-relation mean is 0.5508 (seed range 0.5457--0.5551), above $1/2$. The pooled result therefore does not establish a negative orientation for a uniformly sampled eligible relation. Relation weighting changes the estimand, and eligible relations themselves depend on the fixed predictor's errors. This algebraic accounting locates the reversal without identifying a causal training or observation mechanism.

\begin{table}[t]
\centering\small
\setlength{\tabcolsep}{3pt}
\begin{tabular}{llrrrrrr}\toprule
Predictor & Dataset & Pooled & Same mass (\%) & Same & Cross & Macro & Eligible\\\midrule
DistMult & FB15k-237 & 0.6671 & 1.59 & 0.7769 & 0.6653 & 0.7193 & 128/127/136 \\
DistMult & WN18RR & 0.7412 & 6.53 & 0.6048 & 0.7507 & 0.6981 & 7/7/8 \\
ComplEx & FB15k-237 & 0.6893 & 1.63 & 0.7813 & 0.6878 & 0.7440 & 124/130/120 \\
ComplEx & WN18RR & 0.7680 & 6.57 & 0.6288 & 0.7778 & 0.7185 & 9/7/8 \\
Recip. ComplEx & FB15k-237 & 0.3639 & 1.72 & 0.4554 & 0.3623 & 0.5508 & 137/136/127 \\
Recip. ComplEx & WN18RR & 0.6341 & 9.21 & 0.5300 & 0.6447 & 0.5932 & 8/8/8 \\
\bottomrule\end{tabular}
\caption{Raw missing-role score AUROC decomposed by whether the error and correct query share a relation. Same/Cross are conditional pair AUROCs; same mass is the fraction of all error--correct pairs within a relation. Macro averages only relations containing both classes; Eligible gives their counts for seeds 17/29/43. All values except counts are equal-seed means. The exact identity uses per-seed pair masses and components; multiplying separately averaged columns need not reproduce the pooled column. Full integer counts, contributions, all relations and seeds are archived. These are post-result descriptive estimands, not causal or sampling-inference results.}
\label{tab:relation-pair-diagnostic}
\end{table}

\paragraph{Strong comparator controls in the improved regime.}
All seven LR/HGB feature specifications are refitted on bins 0/2 with their existing hyperparameters, after checkpoint selection and without further search. Base + binary LR still improves mean 85\%-coverage risk over Base LR by 0.240/0.073 percentage points on FB15k-237/WN18RR. Continuous-count LR now misses the hindsight floor in all six runs, with mean excess risks 0.411/0.321 points, although it retains 99.21\%/99.39\% of all correct predictions. Exact original-regime saturation therefore does not transfer, and the remaining high-coverage headroom is still small.

\begingroup\setlength{\tabcolsep}{4pt}
\begin{table}[t]
\centering\small
\begin{tabular}{llrrrrr}\toprule
Dataset & Error head & Risk50 (\%) & Risk85 (\%) & AURC & Excess AURC & Brier\\\midrule
FB15k-237 & Base LR & 26.848 & 48.772 & 0.25914 & 0.06324 & 0.13295\\
FB15k-237 & Base + binary LR & 26.395 & 48.532 & 0.25654 & 0.06065 & 0.13042\\
FB15k-237 & Base + counts LR & 24.026 & 48.224 & 0.24503 & 0.04913 & 0.11695\\
FB15k-237 & Base + counts + binary LR & 24.045 & 48.251 & 0.24463 & 0.04873 & 0.11630\\
FB15k-237 & Base HGB & 25.178 & 48.509 & 0.25109 & 0.05519 & 0.12315\\
FB15k-237 & Base + binary HGB & 24.634 & 48.209 & 0.24738 & 0.05149 & 0.11945\\
FB15k-237 & Base + counts HGB & 23.288 & 48.201 & 0.24196 & 0.04606 & 0.11284\\
\addlinespace
WN18RR & Base LR & 14.906 & 47.644 & 0.20649 & 0.01530 & 0.05293\\
WN18RR & Base + binary LR & 14.534 & 47.571 & 0.20544 & 0.01425 & 0.04993\\
WN18RR & Base + counts LR & 14.385 & 47.425 & 0.20486 & 0.01366 & 0.04989\\
WN18RR & Base + counts + binary LR & 14.236 & 47.338 & 0.20421 & 0.01301 & 0.04779\\
WN18RR & Base HGB & 14.038 & 47.440 & 0.20136 & 0.01017 & 0.04393\\
WN18RR & Base + binary HGB & 13.914 & 47.381 & 0.20075 & 0.00956 & 0.04318\\
WN18RR & Base + counts HGB & 13.591 & 47.367 & 0.20022 & 0.00902 & 0.04231\\
\addlinespace
\bottomrule\end{tabular}
\caption{Predeclared exploratory stronger scorer controls on selected reciprocal ComplEx predictors. All scorer fits use only validation hash bins 0 and 2. Risk is percent; AURC, excess over hindsight oracle AURC, and Brier are unscaled. All methods are retained; means over three optimizer seeds do not provide population inference.}
\label{tab:reciprocal-score-controls}
\end{table}

\endgroup

Adding binary bits to Counts LR improves AURC and Brier in every seed, but worsens mean FB15k-237 risk85 by 0.0267 percentage points and risk50 by 0.0194 points; risk85 worsens in two of its three seeds. On WN18RR its mean risk85 improves by 0.0875 points. Binary and continuous roles both improve mean HGB AURC/Brier over Base HGB, using the same family and hyperparameter budget. Counts HGB has mean excess AURC $0.04606$ on FB15k-237 and $0.00902$ on WN18RR, conditional on each predictor's hindsight floor. It is not uniformly best: its risk85 is worse than binary HGB in two FB15k-237 seeds, and its AURC is worse in one WN18RR seed. Table~\ref{tab:reciprocal-score-controls} retains the raw metrics and both coverages. Cross-predictor raw AURC or Brier changes mix error-prevalence and score-quality changes and cannot isolate an uncertainty-scoring improvement.

\subsection{Does role evidence remain useful beyond explicit ensemble disagreement?}\label{sec:ensembles}
We construct exactly one equally weighted three-member ensemble from seeds 17/29/43 for each dataset and each of DistMult, original ComplEx and reciprocal ComplEx. There are six fixed ensembles, with no member, weight, temperature or aggregation search and no independent ensemble-level repeats. For each query, let $p_m$ be the temperature-one softmax of member $m$'s tail scores after masking train-known tails. Prediction is the highest-probability tail under $\bar p=\tfrac13\sum_m p_m$, breaking ties by sorted training ID. This adapts the probability-mixture convention of deep ensembles \cite{deepensembles}; normalized member scores are not thereby calibrated probabilities of factual truth.

The baseline error head uses ensemble top probability, probability margin, predictive entropy and original endpoint/relation frequencies. A nested version adds three disagreement features: $H(\bar p)-\tfrac13\sum_m H(p_m)$, one minus the largest hard-vote fraction, and the population standard deviation of the three probabilities assigned to the ensemble-chosen tail. The entropy difference is generalized Jensen--Shannon disagreement. Although a corresponding formula occurs in Bayesian acquisition \cite{galactive}, these members are optimization-seed algorithms, not posterior samples. No epistemic/aleatoric decomposition is inferred. The same LR and HGB families then separately add binary roles, continuous role counts, or both. All ten heads fit on validation bins 0/2; bin 1 is recorded without selection and bin 3 is withheld because reciprocal members selected checkpoints there. Features use the ensemble's predicted tail and original training counts.

\begin{table}[t]
\centering\small
\setlength{\tabcolsep}{4pt}
\begin{tabular}{llrrrrr}\toprule
 & & Accuracy & \multicolumn{2}{c}{Binary $\Delta$AURC} & \multicolumn{2}{c}{Counts $\Delta$AURC}\\
Predictor/stage & Dataset & (\%) & LR & HGB & LR & HGB\\\midrule
DistMult & FB15k-237 & 12.95 & -0.037170 & -0.018256 & -0.071931 & -0.032414 \\
DistMult & WN18RR & 40.35 & -0.007561 & -0.002056 & -0.012566 & -0.002781 \\
ComplEx & FB15k-237 & 12.26 & -0.035889 & -0.019902 & -0.070314 & -0.035930 \\
ComplEx & WN18RR & 39.75 & -0.002318 & -0.001298 & -0.002210 & -0.001775 \\
Reciprocal ComplEx & FB15k-237 & 47.30 & -0.001385 & -0.000849 & -0.008947 & -0.004599 \\
Reciprocal ComplEx & WN18RR & 46.15 & +0.000111 & -0.000418 & -0.000438 & -0.000932 \\
\bottomrule\end{tabular}
\caption{Role-feature additions to the same confidence/frequency/disagreement baseline on each fixed three-member ensemble. Negative differences favor the added features. These are six individual ensemble tasks, not means across repeated ensembles or independent graph replications. No sampling intervals are implied.}\label{tab:ensemble-role}
\end{table}

Table~\ref{tab:ensemble-role} shows the role-feature contrasts conditional on each fixed ensemble prediction. Counts improve AURC and Brier in all twelve ensemble/head comparisons. Binary features improve AURC in five of six LR cases and all six HGB cases; their Brier scores improve in all six LR cases and five HGB cases. The improved-regime ensembles reach observed-target accuracies 47.30\%/46.15\% on FB15k-237/WN18RR, but prediction gains and score increments are separate questions. The six tasks and shared members are not independent replications.

The unfavorable reciprocal/WN18RR case distinguishes the metrics. Adding binary roles to its disagreement-aware LR increases AURC by $0.000111$ while reducing Brier. The corresponding HGB addition reduces AURC by $0.000418$, but increases Brier by $0.000222$ and risk50 by 0.1488 percentage points; risk85 is unchanged. Continuous-role HGB improves AURC, Brier and risk50 there, yet worsens risk85 by 0.0875 points. Even the disagreement increment is not uniformly favorable: adding it to confidence/frequency LR in this task increases AURC by $0.000290$ and risk85 by 0.3063 points. These contrasts give no uniformly dominant uncertainty representation.

Conditional on continuous counts, adding binary roles produces identical HGB scores in all six ensembles. For LR it improves Brier throughout, but worsens risk85 for both reciprocal ensembles and slightly worsens reciprocal/WN18RR AURC. The deterministic features can change a restricted fit without adding conditional information. All ten heads, six raw disagreement/confidence scores and both reported coverages are retained in the archived ledger summaries; none was selected by a favorable test result.

\subsection{Prospectively fixed external-dataset check on CoDEx-S}\label{sec:external}
To test whether the two-graph findings extend beyond the data that motivated them, we fixed a CoDEx-S study using the official split of Safavi and Koutra~\cite{codex}. We chose this dataset from published metadata and computational feasibility before fetching its triples or measuring its outcomes. The plan was committed locally before data access; integrity checks were finalized before training. This is a prospectively specified external-dataset check, not a public preregistration, an independently executed replication, or a random sample of graphs. Earlier FB15k-237/WN18RR results informed its hypotheses.

We retained the same seeds, 32-complex-dimensional sampled-training recipe (30 epochs), and eight-epoch reciprocal continuation. The sampled checkpoint is fixed; the continued checkpoint maximizes accuracy on validation hash bin 3, with the earliest epoch selected on ties. Scorer fitting uses bins 0/2 (703 queries), and threshold fitting uses bin 1 (362 queries). Bin 3 contains 350 queries; selected continued-training epochs are 8/4/6. All six model states and all seven fixed LR/HGB heads were saved before the evaluation command first parsed the test outcomes. The downloader had already fetched and hashed opaque test-file bytes for integrity; these were not parsed to construct test queries or outcomes during training. No CoDEx-specific hyperparameter search was performed. A separately recorded relation-identity control, described below, was also fitted and selected before this first outcome inspection.

Of 1,828 positive test triples, grouping yields 1,460 eligible queries. Removing the 355 validation-overlap keys leaves 1,105 test queries with 1,244 recorded targets. There are no unseen-identifier or train-known-target exclusions. Candidates, role counts, and predicted-endpoint features follow the original training-only contract. The published hard-negative annotations are not used: this remains observed-target recovery, not a factual-truth classification task.

The four predefined primary contrasts are Counts minus Base AURC for LR and HGB in each of the two training stages. Our directional consistency criterion requires all four seed-mean differences to be below $-10^{-6}$; it is a descriptive rule, not a significance test. All four satisfy it, with every individual seed also favorable (Table~\ref{tab:external-codex}). Brier improves in all twelve stage/head/seed comparisons as well. At 85\% batch coverage, the mean risk improvements are 1.065/0.390 percentage points for sampled LR/HGB and 0.390/0.461 points for reciprocal LR/HGB. The latter correspond to about 3.67/4.33 additional correct predictions among the same 939 retained predictions; practical gains remain bounded by each model's correct answers. The reciprocal HGB mean conceals two adverse seeds: Counts changes risk85 by $+0.2130/+0.1065/-1.7039$ percentage points (seeds 17/29/43), retaining respectively 2 fewer, 1 fewer, and 16 more correct answers than Base. The primary AURC result is therefore more consistent than this particular operating-point benefit.

\begin{table}[t]
\centering\small
\setlength{\tabcolsep}{4pt}
\begin{tabular}{llrrrr}\toprule
Stage & Head & $\Delta$AURC & Seed range & $\Delta$Brier & $\Delta R_{.85}$ (pp)\\\midrule
Sampled & LR & -0.11424 & [-0.11646, -0.11277] & -0.04019 & -1.065\\
Sampled & HGB & -0.06184 & [-0.06410, -0.06006] & -0.02434 & -0.390\\
Reciprocal & LR & -0.00605 & [-0.00689, -0.00548] & -0.00390 & -0.390\\
Reciprocal & HGB & -0.01050 & [-0.01210, -0.00859] & -0.00707 & -0.461\\
\bottomrule\end{tabular}
\caption{Prospectively fixed CoDEx-S counts-minus-base contrasts on identical predictions, candidates and observed targets. All four predefined comparisons are shown; ranges span three optimizer seeds, not confidence intervals.}
\label{tab:external-codex}
\end{table}

Mean observed-target accuracy rises from 14.36\% to 45.58\%. The raw missing-role score again reverses direction, from mean error AUROC 0.7290 to 0.4297, with each continued-training value below $1/2$. This reproduces the qualitative sign change under the fixed external recipe, but changing the objective, optimization, inverse relations and training budget together still does not identify its cause. All algorithms and seeds share this one graph and observation process.

For each fitted head we additionally set a validation-bin-1 threshold targeting at least 85\% acceptance, retaining the entire boundary tie. The threshold is then held fixed on test queries (Table~\ref{tab:external-codex-threshold}). Counts lower the mean accepted error in these four comparisons, but achieved acceptance differs between heads and falls below 85\% in several cases. These are measured policies, not guarantees of a risk level or improvements at identical online acceptance. Complete per-seed rates and all seven heads are retained in the artifacts.

\begin{table}[t]
\centering\small
\begin{tabular}{llrrrr}\toprule
Stage & Head & Base rate & Counts rate & Base risk & Counts risk\\\midrule
Sampled & LR & 82.78 & 81.72 & 83.77 & 82.45\\
Sampled & HGB & 80.87 & 81.09 & 83.05 & 82.30\\
Reciprocal & LR & 83.74 & 83.68 & 49.06 & 48.62\\
Reciprocal & HGB & 85.79 & 85.13 & 50.18 & 49.40\\
\bottomrule\end{tabular}
\caption{CoDEx-S test acceptance and accepted error (percent), averaged over three seeds, under each head\textquotesingle s validation-bin-1 threshold targeting at least 85\% development acceptance. All boundary ties are accepted. Different achieved test rates preclude interpreting risk differences as equal-coverage benefits.}
\label{tab:external-codex-threshold}
\end{table}

\paragraph{Post-result relation accounting.}
After inspecting these new test results, we repeated the exact pair decomposition from Section~\ref{sec:reciprocal}; this explanation is not a pretest endpoint. CoDEx-S shows the same distinction: mean within-relation pair-weighted AUROC falls from 0.7039 to 0.4833, while equal-relation macro AUROC falls from 0.8196 to 0.5645 and remains above $1/2$ in all three seeds (Table~\ref{tab:relation-pair-external}). Same-relation pairs account for 11.60\% of all error--correct pairs in the improved regime. Thus a second graph exhibits both the pooled reversal and its relation-weighting boundary. Neither decomposition identifies what would happen under an intervention on training support or the graph observation process.

\begin{table}[t]
\centering\small
\setlength{\tabcolsep}{3pt}
\begin{tabular}{lrrrrrr}\toprule
Stage & Pooled & Same mass (\%) & Same & Cross & Macro & Eligible\\\midrule
Sampled & 0.7290 & 9.51 & 0.7039 & 0.7317 & 0.8196 & 23/20/21 \\
Reciprocal & 0.4297 & 11.60 & 0.4833 & 0.4226 & 0.5645 & 25/26/26 \\
\bottomrule\end{tabular}
\caption{CoDEx-S raw-coverage AUROC pair decomposition, added after inspecting the new test results. Same/Cross are conditional error--correct pair AUROCs; Same mass is the within-relation fraction of all pairs. Macro gives equal weight only to relations with both classes; Eligible gives counts for seeds 17/29/43. Other entries are descriptive equal-seed means. Exact per-seed integer contributions and undefined cases are archived. Products of separately averaged columns need not reproduce pooled AUC. This analysis is not part of the external pretest endpoints and makes no causal or population-inference claim.}
\label{tab:relation-pair-external}
\end{table}

A separately authored checker reconstructed all 15,120 development/test query rows from the six saved checkpoints, training triples and target files, and freshly refitted all 42 heads. Predictions, target membership, count features, probabilities, metrics, thresholds and primary contrasts matched, with zero maximum feature/score discrepancy. This checks the saved experiment's internal reproducibility; it is not independent retraining, validation of unsaved intermediate checkpoints, or statistical confirmation of a graph-population effect.

\subsection{Do counts add value after relation identity and matched tuning?}\label{sec:relation-identity}
To distinguish role counts from a relation-specific intercept, we add a one-hot relation identity to both sides of the count comparison. Four LR specifications use Base, Base plus counts, Base plus relation ID, and Base plus both. Base includes the explicit disagreement features for ensembles. Numeric standardizers and the relation vocabulary are fit on the scorer-fitting partition only; unseen selection/test relations receive an all-zero indicator vector. Each specification has the same $C\in\{0.1,1,10\}$ grid and selects minimum AURC on the disjoint development-selection partition, breaking exact ties toward smaller $C$, without refitting on selection outcomes. This is a small matched search, not exhaustive tuning or a claim that equal budgets imply equal effective capacity.

The six reciprocal predictors, six fixed ensembles, and six external predictor stages give 18 paired tasks. The first twelve are exploratory reuse. The CoDEx-S head plan was recorded during training and every head was fit/selected before the first external test evaluation; this is supplemental to the earlier four-contrast primary plan. The supplemental ordering is recorded by the execution sequence; the primary evaluator gate itself checks the seven original heads, not these later relation-ID heads. Table~\ref{tab:relation-identity-groups} reports the counts increment conditional on relation ID. AURC and Brier improve in all 18 tasks, while risk85 improves in 17 and ties one. Risk50 worsens for both reciprocal ensembles, by 0.039/0.074 percentage points on FB15k-237/WN18RR. Gains therefore remain endpoint-dependent.

The reciprocal predictor means shrink to $\Delta$AURC $-0.00471/-0.00224$ on FB15k-237/WN18RR. The corresponding ensemble increments are only $-0.00125/-0.00114$. On external CoDEx-S, means are $-0.09290$ for sampled and $-0.00512$ for reciprocal training. Counts retain a measured increment after this comparator strengthening, but the small improved-regime magnitudes constrain practical claims. Tables~\ref{tab:relation-identity-scores}--\ref{tab:relation-identity-risks} retain every task and all four specifications, including unfavorable differences. Multiple reused tasks and correlated seeds are not independent replications.

\begin{table}[t]
\centering\scriptsize\setlength{\tabcolsep}{3pt}
\begin{tabularx}{\linewidth}{@{}Xrrrrr@{}}\toprule
Group & Runs & $\Delta$AURC & $\Delta$Brier & $\Delta$Risk50 (pp) & $\Delta$Risk85 (pp)\\\midrule
ensemble / ComplEx / FB15k-237 & 1 & -0.040282 & -0.020538 & -2.271845 & -0.331239\\
ensemble / ComplEx / wn18rr & 1 & -0.002415 & -0.010559 & -1.562500 & -0.131291\\
ensemble / DistMult / FB15k-237 & 1 & -0.041300 & -0.022605 & -1.922330 & -0.205597\\
ensemble / DistMult / wn18rr & 1 & -0.004222 & -0.014467 & -2.306548 & -0.131291\\
ensemble / ReciprocalComplEx / FB15k-237 & 1 & -0.001251 & -0.001472 & +0.038835 & -0.125642\\
ensemble / ReciprocalComplEx / wn18rr & 1 & -0.001144 & -0.000643 & +0.074405 & -0.218818\\
external / reciprocal / codex-s & 3 & -0.005117 & -0.003373 & -0.362319 & -0.532481\\
external / sampled / codex-s & 3 & -0.092905 & -0.034364 & -5.978261 & -0.461484\\
reciprocal / ReciprocalComplEx / FB15k-237 & 3 & -0.004710 & -0.005834 & -1.009709 & -0.300781\\
reciprocal / ReciprocalComplEx / wn18rr & 3 & -0.002238 & -0.002848 & -0.694444 & -0.145879\\
\bottomrule\end{tabularx}
\caption{Relation-aware count increments. Negative differences favor counts. Three-seed means are descriptive; each ensemble group contains one fixed ensemble. CoDEx-S is a pretest extension; the original two-graph controls are exploratory. No independence or significance claim.}\label{tab:relation-identity-groups}
\end{table}

\subsection{Changing the prediction policy: training-supported candidates}
After inspecting the primary results, we reran prediction from the six original DistMult checkpoints while restricting each candidate tail to an entity with $n_T(t,r)>0$ in training, still excluding training-known answers. This is an observed-range restriction motivated by type constraints, not an assertion of ground-truth entity types. The development and test query sets are unchanged, including queries for which no recorded target is reachable. If the restricted candidate set is empty, the policy falls back to the original training-only candidate universe without inspecting any target. This occurs for one FB15k-237 development-selection query and no test queries. For a singleton candidate set, the top-two margin is assigned the finite sentinel zero and entropy is zero; 693 FB15k-237 test queries have this case. No queries are dropped.

All feature extraction and scorer fitting are repeated on the same development partitions for the changed predictor. Accordingly, differences from the original candidate rule combine a prediction change with new confidence features and fitted scorers; they cannot be read as isolated abstention gains. The contrast between Base + binary LR and Base LR is made separately within each frozen predictor. This is post-result exploratory analysis, with no change to the recorded primary comparison.

\begin{table}[t]
\centering\small
\begin{tabular}{llrr}
\toprule
Dataset & Candidate policy & Accuracy (\%) & $\Delta$ risk (pp)\\
\midrule
FB15k-237 & Train-only & 10.56 & -0.537 [-0.685, -0.445]\\
FB15k-237 & Observed range & 25.79 & -0.019 [-0.034, 0.000]\\
WN18RR & Train-only & 38.16 & -0.277 [-0.438, -0.175]\\
WN18RR & Observed range & 28.05 & -0.073 [-0.088, -0.044]\\
\bottomrule
\end{tabular}
\caption{Exploratory candidate-policy diagnostic using the same checkpoints and query populations. Accuracy is full-coverage observed-target recovery. $\Delta$ risk is coverage logistic minus confidence-and-count logistic at 85\% coverage within each predictor; entries are seed means [minimum, maximum]. Scorers are refitted on their corresponding development predictions. This changes the predictor and is not an isolated abstention comparison across policies.}
\label{tab:type-constraint}
\end{table}

The restriction increases mean FB15k-237 full-coverage accuracy from 10.56\% to 25.79\% (15.23 percentage points), but decreases WN18RR accuracy from 38.16\% to 28.05\% (10.10 points). Its within-predictor risk reduction for Base + binary LR against Base LR at 85\% coverage shrinks to 0.019 percentage points on FB15k-237 (seed range 0--0.034) and 0.073 points on WN18RR (0.044--0.088). Missing-tail support no longer varies among restricted test predictions by construction. The incremental gains are sensitive to the candidate policy. Because this intervention also changes predictions, error prevalence, and confidence, it cannot quantify how much of the original gain is explained by candidate support or isolate a causal mechanism.

The restriction's cost is visible directly from the target sets: it excludes \emph{all} observed answers for 683 of 10,301 FB15k-237 test queries (6.63\%) and 744 of 2,689 WN18RR queries (27.67\%). These queries remain in all reported denominators. Training absence therefore cannot safely be treated as a hard type rule: a held-out true fact may place an entity in a previously unseen relational role. The WN18RR accuracy loss is an empirical failure case for that assumption, and the FB15k-237 gain does not justify universal support-based filtering.

Paired accounting on the same observed targets shows that the restriction changes incorrect to correct predictions for a mean 15.93\% of all FB15k-237 queries while changing correct to incorrect for 0.71\%; on WN18RR the corresponding fractions are 1.17\% and 11.27\%. Before rounding, their differences exactly account for the accuracy changes on these query sets, without identifying effects on unrecorded truth or a real-world causal mechanism.

\subsection{Sensitivity to grouping the held-out query pool}
Excluding query keys shared by the original validation and test splits alters the population. On FB15k-237, 46.89\% of excluded queries have multiple recorded answers, versus 9.80\% of retained queries; their missing-head-support rates are 3.70\% and 45.06\%, respectively. Table~\ref{tab:grouped-pool-populations} gives corresponding WN18RR comparisons. These differences motivate a separate sensitivity analysis rather than treating the retained population as representative of every benchmark query.

\begin{table}[t]
\centering
\small
\begin{tabular}{llrrrr}
\toprule
Dataset & Query population & $n$ & Multi (\%) & Head gap (\%) & Tail gap (\%) \\
\midrule
FB15k-237 & Primary retained & 10,301 & 9.80 & 45.06 & 6.63 \\
FB15k-237 & Primary excluded & 3,408 & 46.89 & 3.70 & 5.69 \\
FB15k-237 & Pooled test & 11,217 & 23.02 & 39.44 & 5.89 \\
WN18RR & Primary retained & 2,689 & 1.79 & 61.84 & 27.67 \\
WN18RR & Primary excluded & 125 & 16.80 & 8.00 & 56.00 \\
WN18RR & Pooled test & 2,664 & 4.32 & 60.62 & 29.35 \\
\bottomrule
\end{tabular}
\caption{Descriptive population composition. Multi denotes multiple observed answers; head gap denotes no training head--relation support; tail gap denotes every observed answer lacking training tail--relation support. Retained/excluded populations use original test answers; pooled test uses unioned validation/test answers. Held-out answer support is used only in this description, never as a decision-time score.}
\label{tab:grouped-pool-populations}
\end{table}

We union eligible validation and test answers within each $(h,r)$ group, then assign each complete group by SHA256 modulo four: bin zero to fitting, bin one to selection, and bins two/three to testing. Group assignment uses query identifiers only. Training triples, all twelve checkpoints, training-only candidate masks and the nine original scorer definitions are unchanged; fitting and selection are repeated on the new partitions. This removes deterministic exclusion of overlap groups while retaining query separation. The new fit/select/test sizes are 5,617/5,539/11,217 for FB15k-237 and 1,350/1,383/2,664 for WN18RR.

\begin{table}[t]
\centering
\small
\begin{tabular}{llrr}
\toprule
Model & Dataset & Full accuracy (\%) & Coverage $\Delta$ risk (pp) \\
\midrule
DistMult & FB15k-237 & 12.83 [12.40, 13.09] & -1.07 [-1.15, -1.03] \\
DistMult & WN18RR & 36.27 [36.04, 36.64] & -0.19 [-0.40, -0.09] \\
ComplEx & FB15k-237 & 12.08 [11.83, 12.39] & -0.62 [-0.70, -0.53] \\
ComplEx & WN18RR & 35.69 [35.17, 36.19] & -0.46 [-0.49, -0.40] \\
\bottomrule
\end{tabular}
\caption{Post-result grouped-pool sensitivity using all twelve frozen checkpoints. Queries are assigned by SHA256 modulo four after unioning eligible original validation/test answer sets. Entries are seed means [minimum, maximum]. The risk contrast at 85\% batch coverage is coverage logistic minus entity-count logistic. These custom partitions reuse previously examined data and change the query population and target sets; they are not independent confirmation or standard benchmark results.}
\label{tab:grouped-pool-results}
\end{table}

The coverage-versus-baseline risk reduction persists in all twelve runs, with means 1.07/0.19 percentage points for DistMult and 0.62/0.46 for ComplEx on FB15k-237/WN18RR. These initial contrasts use the original confidence-and-count baseline. Pooling both changes which queries are tested and enlarges some observed target sets; the new numbers cannot isolate the effect of overlap exclusion. The study also reuses previously inspected data. It is a custom static-population sensitivity analysis, not a new standard benchmark result or independent confirmation.

\paragraph{Strong scorer controls on the pooled population.}
To cross the population and comparator checks, we subsequently fit the same four LR and three HGB feature specifications to the cached grouped development-fit arrays, without additional tuning. The complete seven-head comparison appears in Table~\ref{tab:grouped-score-controls} (Appendix~\ref{app:empirical-detail}). Counts LR reaches the 85\%-coverage hindsight floor in eleven of twelve pooled runs. Adding the binary indicators improves this endpoint in one run, ties eight, and worsens three; thus exact saturation in every run is specific to the original population. At the group-mean level, binary augmentation improves Counts-LR AURC and Brier in all four model/dataset groups. Both binary-role and continuous-role HGB reduce mean AURC and Brier relative to base HGB in all four groups, under the same family and hyperparameter budget. Counts HGB also attains the 85\% floor in eleven runs; in the remaining ComplEx/WN18RR seed it discards two correct predictions and has 0.0883 percentage points higher risk than binary-role HGB. Thus favorable mean AURC and Brier do not imply dominance at every operating point. These executed results extend the comparator-dependent conclusion to the pooled population, with no independent-confirmation claim.

The target-set change is measurable. Among the pooled test queries, 1,713 FB15k-237 and 62 WN18RR keys occurred in both original held-out splits; every such group's union adds answers beyond its original test set. On FB15k-237, respectively 184/201/200 DistMult predictions and 192/170/178 ComplEx predictions (seeds 17/29/43) are correct only because an added validation answer is accepted. The WN18RR counts are 7/10/7 and 12/15/10. These counts use the overlap denominators 1,713 and 62, not all 11,217 and 2,664 pooled test queries; the equal-seed mean overlap fractions are 11.38\%/10.51\% and 12.90\%/19.89\% for DistMult/ComplEx. This is same-prediction label accounting within overlap groups. It cannot attribute the overall pooled-versus-original accuracy change, whose query composition also changes.

\subsection{Crossing predictor regime, grouped population, and label policy}\label{sec:reciprocal-grouped}
A further post-outcome sensitivity reuses all six selected reciprocal ComplEx checkpoints on a grouped held-out pool. Eligible original validation and test targets are unioned within each query. Original-identifier SHA256 modulo four assigns whole queries to fitting bin0, threshold bin1, evaluation bin2, and discarded bin3. Unlike the original-predictor pooled study, bin3 cannot be evaluated here: its original validation outcomes selected the reciprocal checkpoints. All bin3 queries are therefore excluded before prediction or scoring. The original reciprocal error heads also cannot be reused because their fitting partition included bin2; all seven LR/HGB heads and their standardizers are instead refitted on pooled bin0 alone, with the same fixed hyperparameters. No link predictor is retrained. Candidates and counts remain defined by original training triples only.

The fit/threshold/test sizes are 5,617/5,539/5,568 on FB15k-237 and 1,350/1,383/1,368 on WN18RR; 5,649 and 1,296 bin3 queries are discarded. This design crosses population and predictor-regime checks without reusing checkpoint-selection queries for performance evaluation. It still follows earlier outcome inspection and is neither independent confirmation nor a causal analysis.

\begin{table}[t]
\centering\small
\setlength{\tabcolsep}{4pt}
\begin{tabular}{llrrrr}\toprule
Dataset & Error head & Risk50 (\%) & Risk85 (\%) & AURC & Brier\\\midrule
FB15k-237 & Base LR & 29.646 & 49.669 & 0.274605 & 0.146016\\
FB15k-237 & Base + binary LR & 29.442 & 49.310 & 0.271752 & 0.143528\\
FB15k-237 & Base + counts LR & 26.042 & 48.556 & 0.255111 & 0.126583\\
FB15k-237 & Base + counts + binary LR & 25.635 & 48.528 & 0.254319 & 0.125545\\
FB15k-237 & Base HGB & 26.724 & 48.972 & 0.260887 & 0.131959\\
FB15k-237 & Base + binary HGB & 26.365 & 48.697 & 0.257798 & 0.129031\\
FB15k-237 & Base + counts HGB & 24.569 & 48.429 & 0.248867 & 0.118854\\
\addlinespace
WN18RR & Base LR & 16.423 & 48.021 & 0.211007 & 0.057368\\
WN18RR & Base + binary LR & 16.228 & 48.049 & 0.210165 & 0.054859\\
WN18RR & Base + counts LR & 16.131 & 48.049 & 0.209360 & 0.054780\\
WN18RR & Base + counts + binary LR & 15.887 & 47.935 & 0.208874 & 0.052624\\
WN18RR & Base HGB & 15.351 & 47.762 & 0.205458 & 0.047814\\
WN18RR & Base + binary HGB & 15.302 & 47.705 & 0.204674 & 0.046606\\
WN18RR & Base + counts HGB & 15.253 & 47.734 & 0.204352 & 0.046499\\
\addlinespace
\bottomrule\end{tabular}
\caption{Post-outcome grouped-population sensitivity for six selected reciprocal ComplEx checkpoints. All seven heads are refitted on pooled query hash bin0; bin1 supplies fixed thresholds; only bin2 is evaluated, excluding checkpoint-selection bin3. Entries are means over three seeds, not independent replications or confidence intervals. Base includes confidence and entity/relation frequencies; counts adds continuous role counts.}
\label{tab:reciprocal-grouped-controls}
\end{table}

Mean observed-target accuracy is 44.09\%/44.64\% on FB15k-237/WN18RR. The raw scalar missing-role AUROC is 0.3821/0.6211; the reversed FB15k-237 orientation persists in all three seeds. Relative to each same-family Base head, both binary and continuous role additions improve AURC and Brier in every run. Continuous-count additions reduce mean LR/HGB AURC by 0.01949/0.01202 on FB15k-237 and 0.00165/0.00111 on WN18RR. These favorable curve and probability-score comparisons do not imply improvement at every coverage. On WN18RR, both Base + binary LR and Base + counts LR have 0.0287 percentage points \emph{higher} mean risk85 than Base LR. Binary LR worsens one seed and ties two; Counts LR worsens one and improves two. Counts HGB also has 0.0287 points higher mean risk85 than binary HGB, and its Brier score is worse than binary HGB in two WN18RR seeds despite a favorable mean difference. Adding binary indicators to Counts LR improves mean AURC and Brier on both datasets.

\paragraph{Separating target enlargement from the changed population.}
Among evaluated bin2 queries, 3,381 FB15k-237 and 721 WN18RR queries have eligible original-test targets; 841 and 30 also have original-validation targets. On these fixed subsets, we hold predictions, features, all seven fitted scores, and threshold policies unchanged, and evaluate both union-target and original-test-target labels. An added validation answer makes 122/142/160 FB15k-237 predictions and 8/5/8 WN18RR predictions correct (seeds 17/29/43). The corresponding mean accuracies change from 38.29\% to 42.47\% and from 42.90\% to 43.87\%.

\begin{table}[t]
\centering\small
\setlength{\tabcolsep}{4pt}
\begin{tabular}{llrrrr}\toprule
Dataset & Targets & Accuracy (\%) & LR $\Delta$AURC & HGB $\Delta$AURC & HGB $\Delta$Brier\\\midrule
FB15k-237 & Union & 42.473 & -0.023146 & -0.014119 & -0.013911\\
FB15k-237 & Original test & 38.292 & -0.015372 & -0.009664 & -0.007345\\
WN18RR & Union & 43.874 & -0.001847 & -0.001445 & -0.001524\\
WN18RR & Original test & 42.903 & -0.001397 & -0.000237 & -0.001153\\
\bottomrule\end{tabular}
\caption{Label sensitivity on the same predictions and scores for pooled bin2 queries with original-test targets: 3,381 FB15k-237 and 721 WN18RR queries. Differences are Counts minus Base within each family; negative favors Counts. Union versus original-test labels are compared on identical subsets. Means can hide adverse seed-level cases.}
\label{tab:reciprocal-grouped-labels}
\end{table}

The label comparison can change scorer conclusions as well as accuracy. Under original-test-only labels, Counts HGB has higher AURC than Base HGB in WN18RR seeds 29 and 43, by 0.000254 and 0.000098; its Brier score is also higher in seed43 by 0.000096. The same two AURC contrasts favor Counts under union labels. Table~\ref{tab:reciprocal-grouped-labels} retains the means, which remain favorable and hide these individual failures. This is label accounting for fixed predictions and scorers trained on pooled union labels. It is not a refitted original-test-label experiment, and it cannot explain the full pooled-versus-original difference, which also changes query composition, fitting data, and evaluation size.

\paragraph{Fixed thresholds on the stronger grouped predictor.}
Each head's threshold gives exactly 85\% expected acceptance on bin1 and is then transferred unchanged to bin2. Table~\ref{tab:reciprocal-grouped-thresholds} reports both achieved coverage and the ratio of expected error to acceptance counts. On WN18RR, Base HGB answers a mean 83.75\% with error ratio 47.04\%, whereas Counts HGB answers 85.45\% with error ratio 48.04\%. Counts HGB answers more queries and has a higher error ratio in every seed, despite its lower mean batch risk85. Conversely, on FB15k-237 its lower error ratio accompanies fewer answers than Base HGB. These unequal operating points do not identify improvement at equal coverage. Together, the grouped results preserve useful role-feature increments while exposing dependence on labels, metrics, and transferred acceptance policies.

\begin{table}[t]
\centering\small
\setlength{\tabcolsep}{4pt}
\begin{tabular}{llrrrr}\toprule
Dataset & Family & \shortstack{Base\\coverage (\%)} & \shortstack{Base\\error (\%)} & \shortstack{Counts\\coverage (\%)} & \shortstack{Counts\\error (\%)}\\\midrule
FB15k-237 & LR & 84.579 & 49.490 & 84.686 & 48.402\\
FB15k-237 & HGB & 85.081 & 49.016 & 84.255 & 48.030\\
WN18RR & LR & 84.917 & 48.091 & 84.893 & 47.929\\
WN18RR & HGB & 83.748 & 47.045 & 85.448 & 48.044\\
\bottomrule\end{tabular}
\caption{Pooled-bin1 thresholds transferred unchanged to bin2 for selected reciprocal checkpoints. Coverage and error ratios are equal-seed means; each error ratio is expected errors divided by expected acceptances, not the expected realized finite-batch risk. Methods have unequal achieved coverages, so these are not equal-coverage performance contrasts.}
\label{tab:reciprocal-grouped-thresholds}
\end{table}

\subsection{Changing the acceptance policy: development-fixed thresholds}
The primary endpoint controls the number of answers within each test batch. To examine transfer of an acceptance rule, a further post-result diagnostic fixes a threshold separately for each frozen DistMult score using development-selection data. It accepts all scores below the threshold and a fitted fraction of its tied boundary, giving exactly 85\% expected development acceptance. The threshold and boundary fraction are then unchanged on test. For selected mixtures this development partition was already used to choose the mixture weight, so it is not a fresh calibration sample. No test outcome chooses a threshold, and no finite-sample coverage guarantee is asserted.

For test acceptance probabilities $a_i$, we report $n^{-1}\sum_i a_i$ and $\sum_i a_i\ell_i/\sum_i a_i$. The second quantity is a ratio of expected error and acceptance counts; it is not the expectation of a realized finite-batch ratio. On FB15k-237, the confidence-and-count baseline attains mean coverage 87.41\% with error ratio 88.40\%, whereas coverage augmentation attains 82.86\% and 87.33\%. The lower error therefore accompanies 4.55 percentage points fewer answers. On WN18RR, the respective pairs are 82.97\%/55.18\% and 83.67\%/55.30\%; the augmented error ratio is higher on two of three seeds. These policies operate at different achieved coverages. They do not establish deployment improvement from the favorable batch-budget comparison.

\paragraph{Reproducibility and remaining scope.}
The archived protocols and configurations record the tasks, exclusions, seeds, objectives, feature sets, and contrasts. A pinned data mirror and per-file hashes identify the downloaded split bytes. Checkpoints, fitting/selection arrays, query-level targets and predictions, feature and score arrays, source hashes, dependency versions, and metrics support re-evaluation. They do not recover the missing provenance of the historical results. Independent verification replays every stored query prediction for the six reciprocal checkpoints. Ensemble verification independently refits all 60 error heads and checks every stored population row; checkpoint prediction replay uses a fixed sample rather than every ensemble query. These engineering checks do not supply independent scientific replication. The measured role-feature effects remain conditional on predictor regime, scorer and metric. None of these static observed-target studies validates temporal generalization, truth verification, calibrated epistemic uncertainty, universal mixture gains, or state-of-the-art KG performance.

\subsection{Relation concentration and fixed-weight comparisons}\label{sec:relation-weights}
The native within-relation AUROC is $\sum_r\pi_r A_r$, where $A_r$ is relation $r$'s tie-aware AUROC and $\pi_r=E_rC_r/\sum_jE_jC_j$. Its eligibility and weights change when predictions change: a relation must have both errors and correct predictions to receive positive mass. Equal-relation macro AUROC is a different estimand. The following accounting was added after reviewer inspection of the earlier decomposition. It is explanatory reuse, including for CoDEx-S, and was not a pretest endpoint.

\paragraph{Concentration and influence.}
The highest-weight reciprocal relation is \texttt{/people/person/profession} on FB15k-237 (874 queries; 418--422 errors), \texttt{\_hypernym} on WN18RR (1,065 queries; 949--960 errors), and P106 on CoDEx-S (334 queries; 218--226 errors). Across seeds, these relations carry 41.48--43.05\%, 63.24--65.07\%, and 70.43--71.63\% of within-relation pair mass, respectively. The leading four carry 73.59--75.27\%, 98.53--98.78\%, and 91.28--91.81\%. The dominant relation's own AUROC is 0.4364--0.4590, 0.5252--0.5516, and exactly 0.5, respectively. High pair mass therefore need not itself drive a negative direction: CoDEx-S P106 dilutes the below-chance contribution of other relations.

For each reciprocal seed, rank relations by $E_rC_r$ (ties by relation name), remove the leading one or four, and renormalize the remaining native within-pair weights. Removing the leading four leaves FB15k-237 AUROCs 0.4767/0.4624/0.4788 and CoDEx-S 0.4213/0.4157/0.4113. WN18RR gives 0.6954/0.6416/0.6046 on only 1.22--1.47\% of original within-pair mass. These are influence diagnostics on fixed predictions, not causal graph interventions. In FB15k-237 seed17, 74 eligible relations are above chance, 34 below, and 29 tied at chance, despite negative pair-weighted AUROC. Every relation's denominators, direction and mass, and both omission sizes, are retained in the result artifact.

\paragraph{Holding eligibility and weights fixed across stages.}
For each paired sampled/reciprocal seed, restrict to relations with both classes in both stages. FB15k-237 retains 96--107 relations and 7,830--8,851 of 10,301 queries; WN18RR retains 7--8 and 2,498--2,677 of 2,689; CoDEx-S retains 19--22 and 987--1,046 of 1,105. On this common set, report all four fixed weighting choices: sampled $E_rC_r$, reciprocal $E_rC_r$, query count $n_r$, and equal relation weights. Each choice is held identical between stages. The two pair choices use outcomes of their respective predictor; neither is a distribution-free target population, and no weight choice is selected for its sign.

\begin{table}[t]
\centering\small
\setlength{\tabcolsep}{4pt}
\begin{tabular}{@{}llrr@{}}\toprule
Graph & Fixed common-set weights & Sampled & Reciprocal mean [min, max]\\\midrule
FB15k-237 & Sampled pairs & 0.7806 & 0.4645 [0.4558, 0.4805] \\
FB15k-237 & Reciprocal pairs & 0.8037 & 0.4544 [0.4477, 0.4581] \\
FB15k-237 & Queries & 0.7763 & 0.4912 [0.4767, 0.5101] \\
FB15k-237 & Equal relations & 0.7359 & 0.5416 [0.5278, 0.5504] \\
wn18rr & Sampled pairs & 0.6288 & 0.5184 [0.4999, 0.5283] \\
wn18rr & Reciprocal pairs & 0.6905 & 0.5301 [0.5267, 0.5335] \\
wn18rr & Queries & 0.6825 & 0.5414 [0.5349, 0.5457] \\
wn18rr & Equal relations & 0.7354 & 0.5985 [0.5947, 0.6025] \\
CoDEx-S & Sampled pairs & 0.7039 & 0.4866 [0.4802, 0.4948] \\
CoDEx-S & Reciprocal pairs & 0.7047 & 0.4831 [0.4799, 0.4860] \\
CoDEx-S & Queries & 0.7588 & 0.4732 [0.4720, 0.4753] \\
CoDEx-S & Equal relations & 0.8146 & 0.5222 [0.5075, 0.5442] \\
\bottomrule\end{tabular}
\caption{Post-review common-eligibility sensitivity. Pair weights are proportional to the stated stage's $E_rC_r$; query weights to $n_r$. Each row holds weights fixed between stages; all four choices are reported. Seed ranges are descriptive.}
\label{tab:relation-fixed-weights}
\end{table}

For both FB15k-237 and CoDEx-S, reciprocal AUROC remains below $1/2$ in every seed under either fixed pair weighting. Common-set equal weights remain above $1/2$ in every seed. Query weighting reveals an additional boundary: FB15k-237 gives 0.5101/0.4767/0.4867, so one seed is above chance, whereas CoDEx-S gives 0.4720/0.4725/0.4753. Even WN18RR changes sides narrowly in one case: sampled-pair weights give 0.499894 for reciprocal seed17. These full choices separate changing aggregation weights from within-relation score changes without furnishing a training mechanism, a uniformly negative relation, or an independently sampled graph effect.

\section{Supporting information and decision arguments}\label{app:theory}
These arguments clarify the historical motivation and the assumptions required to connect information to idealized decisions. They are not guarantees for the operational estimators evaluated in the main text.

\subsection{Historical entity-variance score}
For a diagonal Gaussian entity representation with variance vector $v_e$, an entity-only score is
\begin{equation}
 U_E(h,r,t)=\frac{1}{2d}\sum_{j=1}^d(v_{h,j}+v_{t,j}).
\end{equation}
This is a score constructed from embedding spread. It is not automatically calibrated epistemic uncertainty. Similarly, inverse frequency is a separate count baseline, not an interchangeable measurement of learned variance. We use \emph{entity uncertainty} and \emph{structural novelty} as operational terms; their combination is not a law-of-total-variance decomposition into identifiable uncertainty sources.

\subsection{What entity-only scores can and cannot identify}
Let $P_0$ be a reference-query distribution and $P_1$ an alternative-query distribution. Let $Z=z(h,t)$ collect all features available to an entity-only score, and let $Q_i$ be the induced distribution of $Z$ under $P_i$. Larger scores indicate greater novelty. We use the tie-aware definition
\begin{equation}
 \auc(U)=\Prb(U(X_1)>U(X_0))+\tfrac12\Prb(U(X_1)=U(X_0)),
\end{equation}
with independent $X_i\sim P_i$.

\begin{proposition}[Sharp distributional indistinguishability]\label{prop:tv}
Let $\varepsilon=\tv(Q_1,Q_0)=\sup_A|Q_1(A)-Q_0(A)|$. For every measurable score $U=f(Z)$,
\begin{equation}
 \left|\auc(U)-\tfrac12\right|
 \leq\varepsilon-\frac{\varepsilon^2}{2}\leq\varepsilon.
\end{equation}
The first bound is attainable for every $\varepsilon\in[0,1]$ when the feature distributions and score can vary.
\end{proposition}
\begin{proof}
For $0<\varepsilon<1$, dominate both distributions by $Q_0+Q_1$. The pointwise minimum of their densities defines a common measure of mass $1-\varepsilon$. Normalizing this measure and the two remainders gives probability measures $M,A,B$ with
\[
Q_0=(1-\varepsilon)M+\varepsilon A,\qquad
Q_1=(1-\varepsilon)M+\varepsilon B.
\]
Generate the two independent feature draws through these mixtures. With probability $(1-\varepsilon)^2$, both draws come from $M$; the conditional expected tie-aware comparison is $1/2$ by symmetry. On the remaining event, the comparison lies in $[0,1]$. Thus
\[
\tfrac12(1-\varepsilon)^2\leq\auc(U)
\leq1-\tfrac12(1-\varepsilon)^2,
\]
which is the claimed bound. For $\varepsilon=0$, identical feature distributions give AUROC $1/2$ directly. For $\varepsilon=1$, the result follows from $0\leq\auc(U)\leq1$.
\end{proof}

\paragraph{Tightness and scope.}
Let $\delta_z$ denote a point mass at $z$. On three ordered feature values, take
\[
Q_0=\varepsilon\delta_{-1}+(1-\varepsilon)\delta_0,\qquad
Q_1=(1-\varepsilon)\delta_0+\varepsilon\delta_1,\qquad f(z)=z.
\]
Their total variation is $\varepsilon$. Every comparison outside the common--common event is correctly ordered, giving AUROC $1/2+\varepsilon-\varepsilon^2/2$; reversing $f$ attains the lower endpoint. This elementary common-measure argument refines the weaker total-variation bound without a claim of statistical novelty. It concerns the fixed endpoint features $Z=z(h,t)$; it is not a guarantee for features evaluated at a relation-dependent predicted endpoint. Exact finite-distribution checks verify three-point attainment and randomized cases, but do not certify a benchmark population's total variation.

\begin{corollary}[Exact entity-pair matching]
If the two query populations have the same distribution over $(h,t)$ and differ only in relation assignments, every entity-only score has AUROC $1/2$.
\end{corollary}
The corollary describes a controlled diagnostic, not an assumption that all natural distribution shifts satisfy. It also does not imply that the full triple-scoring model is unable to learn relation-specific behavior.

\paragraph{Why support overlap is insufficient.}
Suppose $Z\in\{0,1\}$, $Q_0(Z=1)=0.01$, and $Q_1(Z=1)=0.99$. Every alternative point has an exactly matching reference feature value, yet $U=Z$ has AUROC $0.99$. Existence of a counterpart does not match probability mass. The original manuscript's nearest-frequency-match rate therefore cannot establish the premise of Proposition~\ref{prop:tv}. A monotone ordering between frequency and variance also does not imply that all equal-frequency entities have equal variance.

\paragraph{A defensible empirical diagnostic.}
For a controlled experiment, sample reference and novel queries with identical empirical entity-pair distributions and compare scores within that design. If exact matching is infeasible, report sample retention, balance, joint-feature discrepancies, and sensitivity to matching. A small estimated discrepancy or failure to reject a two-sample test is not, by itself, a certified upper bound on population total variation.

\subsection{From context to an independently defined error outcome}
The previous bound concerns novelty labels. A separate argument connects context to reliability. Let $L\in\{0,1\}$ be the error of a \emph{fixed} predictor under a fixed evaluation rule, let $Z$ contain entity-only features, and let $R$ contain additional relational context. Define the population error probabilities
\[
 \eta_Z=\E[L\mid Z],\qquad \eta_{ZR}=\E[L\mid Z,R].
\]
\begin{proposition}[Value of context for the Bayes Brier risk]
For any joint distribution of $(L,Z,R)$,
\begin{equation}\label{eq:brier-context}
 \E[(L-\eta_Z)^2]-\E[(L-\eta_{ZR})^2]
 =\E[(\eta_{ZR}-\eta_Z)^2]\geq0.
\end{equation}
The improvement is strict exactly when the conditional error probability varies with $R$ beyond $Z$ on a set of positive probability.
\end{proposition}
\begin{proof}
Write $L-\eta_Z=(L-\eta_{ZR})+(\eta_{ZR}-\eta_Z)$ and expand the square. The cross term has expectation zero since $\E[L-\eta_{ZR}\mid Z,R]=0$. Subtract the remaining first square.
\end{proof}
This is an application of conditional-expectation projection, not a new statistical identity. It establishes potential value of context for a named error outcome without asserting that coverage estimates the Bayes probability. A fitted model can perform worse because of estimation error or shift; calibration and test data are still necessary. Discriminative uncertainty learning has likewise formalized selection through conditional prediction risk \cite{franc}.

\begin{example}[Context-sensitive errors with identical entity features]
Suppose all queries share $Z$, two contexts occur equally often, and their fixed-predictor error probabilities are $0.1$ and $0.9$. Then $\eta_Z=0.5$, while $\eta_{ZR}$ equals $0.1$ or $0.9$. The entity-only Bayes Brier risk is $0.25$ and the context-aware risk is $0.09$, an improvement of $0.16$. Keeping the lower-risk context gives risk $0.1$ at coverage $0.5$; selection using only constant entity features and independent randomization gives risk $0.5$. These are exact properties of a constructed distribution, not KG benchmark measurements.
\end{example}

\subsection{Coverage augmentation and conditional complementarity}
Transform $U_E$ to $S\in[0,1]$ using a mapping fitted on a held-out calibration set; the mapping and score orientation must be fixed before test evaluation. Let $C=U_C$ and define
\begin{equation}\label{eq:mix}
 U_\alpha=\alpha S+(1-\alpha)C,\qquad 0\leq\alpha\leq1.
\end{equation}
The original acronym CAGP is retained only for the original implementation. A Gaussian entity embedding is not sufficient evidence of a Gaussian-process construction; the revised generic method is \emph{coverage-augmented scoring}.

\begin{proposition}[A sufficient condition for complementary ranking]\label{prop:mix}
Let the alternative distribution mix rare-entity queries $E$ and novel-context queries $N$, both with positive probability. Suppose, almost surely for independent class draws:
\begin{enumerate}
\item reference queries have $C(X_0)=0$;
\item novel queries have $C(X_N)\geq m>0$;
\item rare-entity queries satisfy $S(X_E)>S(X_0)$ and $C(X_E)\geq0$;
\item $S(X_0)-S(X_N)\leq D$ for a finite $D\geq0$.
\end{enumerate}
For every $0<\alpha<m/(m+D)$, $U_\alpha$ has AUROC one. It strictly improves on both individual scores only if both individual scores have AUROC less than one on the mixture.
\end{proposition}
\begin{proof}
For a rare-entity/reference pair, the semantic difference is strictly positive and the structural difference is nonnegative. For a novel/reference pair,
\[
 U_\alpha(X_N)-U_\alpha(X_0)\geq-\alpha D+(1-\alpha)m>0.
\]
Thus every alternative/reference comparison is correctly ordered. The final claim follows only when neither individual score already achieves perfect ranking.
\end{proof}
These are strong sufficient conditions, particularly complete rare-entity semantic separation. They are not verified by a high but imperfect empirical AUROC. In the absence of these conditions, mixing is an empirical model-selection decision.

\paragraph{Counterexample to arbitrary mixing.}
Let reference queries have $(S,C)=(0.6,0)$, rare queries $(1,0)$, and novel queries $(0,1)$, with equal mass on the two alternative types. Semantic AUROC is $0.5$ and coverage AUROC is $0.75$. A mixture with $\alpha=0.5$ achieves one, but $\alpha=0.9$ achieves $0.5$, below coverage alone. The existence of a useful weight does not imply that every convex mixture helps.

\paragraph{Coverage ties.}
If all reference queries have coverage score zero and a fraction $\rho$ of rare queries also have score zero, then
\begin{equation}
 \auc(C;E,P_0)=(1-\rho)+\tfrac12\rho=1-\frac{\rho}{2}.
\end{equation}
The original proof omitted the tie term. For $\rho=0.41$, the corrected value is $0.795$, rather than $0.59$.

\subsection{Value of context for optimal selection}
The Brier identity does not by itself establish a gain at a particular coverage. Let $\mathcal F\subseteq\mathcal G$ denote two information sets available before the outcome is revealed. For $0<q\leq1$, define the optimal population selective risk
\[
 R^*_{\mathcal F}(q)=\inf_{a}\frac{\E[La]}q,
 \qquad 0\leq a\leq1,\quad a\text{ is }\mathcal F\text{-measurable},\quad \E[a]=q.
\]
Here $a$ is an acceptance probability, implemented with independent randomization. Allowing fractional acceptance handles atoms and tied scores.
\begin{proposition}[Nested information and selective risk]
For every $q$, $R^*_{\mathcal G}(q)\leq R^*_{\mathcal F}(q)$. An optimal rule accepts the smallest values of $\eta_{\mathcal F}=\E[L\mid\mathcal F]$, randomizing on the boundary as needed to attain coverage $q$.
\end{proposition}
\begin{proof}
Every $\mathcal F$-measurable rule is also $\mathcal G$-measurable, proving the inequality by inclusion of feasible sets. Choose a $q$-quantile $c$ of $\eta_{\mathcal F}$ and let $a^*$ accept below $c$, reject above it, and randomize at $c$. For any feasible $a$, $(\eta_{\mathcal F}-c)(a-a^*)\geq0$ pointwise. Taking expectations and using equal coverage yields $\E[La]\geq\E[La^*]$.
\end{proof}
This standard conditional-risk argument \cite{franc} describes an oracle optimum, not a guarantee for the proposed mixture. Strict Brier improvement need not imply strict selective improvement at every coverage: at $q=1$ all rules have the same risk. The practical experiment must test learned scores at fixed coverages, with context restricted to decision-time information.

\subsection{From estimated error probabilities to selection regret}
Let $\eta=\E[L\mid\mathcal G]$ and let $s\in[0,1]$ be a fitted, $\mathcal G$-measurable estimate of this probability, treated as fixed after development. Let $a^*$ accept the lowest $\eta$ values and $a_s$ the lowest $s$ values, with both rules having population coverage $q$. Put $\delta=|s-\eta|$ and $v=\E[(s-\eta)^2]$.
\begin{proposition}[Plug-in selection regret]
If $\delta\leq\epsilon$ almost surely, then
\begin{align}
0\leq R(a_s)-R(a^*)
&\leq \min\left\{1,\frac{\E[\delta|a_s-a^*|]}q\right\}\\
&\leq \min\left\{1,\frac{2\epsilon\min(q,1-q)}q,
\frac{\E[\delta]}q,\frac{\sqrt{2\min(q,1-q)v}}q\right\}.
\end{align}
\end{proposition}
\begin{proof}
Optimality under the surrogate score gives $\E[s(a_s-a^*)]\leq0$. Subtract this term from $q\{R(a_s)-R(a^*)\}=\E[\eta(a_s-a^*)]$ and bound the remainder by absolute values. Since both acceptance probabilities lie in $[0,1]$ with mean $q$, $\E|a_s-a^*|\leq2\min(q,1-q)$ and $|a_s-a^*|^2\leq|a_s-a^*|$. The other bounds follow from the uniform error bound, $|a_s-a^*|\leq1$, and Cauchy--Schwarz. The cap of one bounds the regret because both risks lie in $[0,1]$; it does not bound the uncapped intermediate expectation.
\end{proof}
The $L^1$ and $L^2$ bounds do not require the uniform approximation assumption; that assumption is used only for the term involving $\epsilon$. This elementary plug-in argument is an adaptation of conditional-risk selection, not a novelty claim. The quantity $v$ equals the excess population Brier risk over $\eta$, by conditional-expectation orthogonality. It is not the raw Brier score or a calibration error. A constant score $s=0.5$ is calibrated in the equal-context example above, but cannot rank the two contexts and has $v=0.16$. Thus calibration alone does not establish selective usefulness. General distinctions between probability estimation, calibration, and downstream decision loss are also studied by Perez-Lebel et al.\ \cite{decisionrisk}.

The bound assumes equal \emph{population} coverage and a fixed fitted score. It does not certify a finite-sample threshold selected on development data, transfer across a changed test distribution, or a novelty score with no probability interpretation. Numerical checks enumerate fractional acceptance on 1,000 random eight-atom distributions, with one coverage sampled from $\{0.1,0.25,0.5,0.85,1\}$ in each trial; no violation above $10^{-12}$ was observed. These are mathematical diagnostics, not evidence of KG performance.

\section{Diagnostic labels, hidden-answer leakage, and proposed extensions}\label{app:diagnostics}
\subsection{Controlled support diagnostics}
The original partition labels a query rare if $\min\{f(h),f(t)\}<\tau$, novel if both entities are frequent but pooled coverage is missing, and reference otherwise. To keep its convention distinct from our role-specific extension, define
\[
c_{\rm pool}(e,r)=\ind\{n_H(e,r)+n_T(e,r)>0\},\qquad
C_{\rm pool}(h,r,t)=1-\tfrac12\{c_{\rm pool}(h,r)+c_{\rm pool}(t,r)\}.
\]
Pooled coverage separates the original novel and reference categories by construction. The rule
\begin{equation}
 B_{\rm pool}(x)=\ind\{\min(f(h),f(t))<\tau\ \text{or}\ C_{\rm pool}(x)>0\}
\end{equation}
exactly recovers that rare-or-novel label whenever its counts, threshold and coverage convention are applied consistently. Replacing $C_{\rm pool}$ by $U_C$ instead reconstructs a \emph{different}, role-separated partition, which is the one implemented by our reference diagnostic helper. The two partitions can disagree: training on $(a,r,b)$ covers both entities for pooled relation $r$, but the reversed query $(b,r,a)$ lacks both required roles. With $\tau=1$, it is pooled-reference and role-specific-novel. Neither reconstruction is an independent prediction-failure test. A model evaluated only on these categories is reconstructing a deterministic label rule, useful for implementation verification but insufficient to establish calibrated uncertainty or predictive utility.

Report the label-reconstruction rule, frequency-only, coverage-only, their fixed average, and the learned mixture on the same partition. Label all such results \emph{support diagnostics}. Static frequency partitions should not be called timestamp-based temporal evaluation.

\paragraph{Training-only counts can still leak the answer.}
The executable diagnostic uses training triples $(h_1,r,c)$, $(h_2,r,c)$, $(d,r,a)$ and $(d,s,b)$, so all entities are seen. Two distinct tail queries $(h_1,r,?)$ and $(h_2,r,?)$ have recorded targets $a$ and $b$, respectively, and equal population weight. A stipulated frozen predictor returns $a$ for both. Both targets and predictions remain eligible after excluding training-known tails for each query; in particular, $a$ is supported for relation $r$ elsewhere, but is not a known answer to either test query. Scoring the true endpoint gives coverage scores $0$ and $1/2$, perfectly detecting the observed-target error. Scoring the predicted endpoint gives $0$ in both cases. The oracle and operational AUROCs are therefore $1$ and $1/2$; risks at half coverage are $0$ and $1/2$. These values are computed by exact enumeration, with no model training. The construction demonstrates possible leakage under a train-only candidate mask, not that it occurred in the historical experiments.

\subsection{Proposed extension: temporal prediction reliability}
The static experiments in the main text do not execute a chronological design. For a future study on a timestamped graph, choose chronological training, calibration, and test cutoffs before examining test outcomes. Compute all coverage and counts from training only. Train the base link predictor on the training window; choose its hyperparameters and the mixture weight using permitted development data; freeze both before the test window. Future queries can contain previously covered and uncovered configurations, so chronological membership is not defined by coverage.

Evaluate error detection using actual prediction outcomes, with the candidate set, filtering rules, and task stated explicitly. Link-ranking errors, triple-classification errors, and OOD membership are distinct targets. A sampled corruption is a candidate negative, not verified falsehood; report negative sampling and false-negative controls. Separate unseen entities from rare but seen entities, because a transductive embedding lookup cannot evaluate a truly new entity without a specified initialization or inductive mechanism.

For top-1 link prediction, let $\ell_i=\ind\{\widehat y_i\neq y_i\}$. At retained fraction $q$, report
\begin{equation}
 R(q)=\frac{\sum_{i\in A_q}\ell_i}{|A_q|},
\end{equation}
where $A_q$ contains the $q$ fraction with lowest uncertainty under a documented tie rule. Compare risk at prespecified coverages and area under the risk--coverage curve. Keep the base predictor fixed when assessing the benefit of abstention.

\subsection{Further comparisons beyond the executed study}
The following comparisons remain proposed extensions. The executed baseline sets and measurements are specified in Section~\ref{sec:new-experiments}; they include fixed algorithmic ensembles, but not posterior embedding inference or chronological evaluation. Further work could compare dedicated probabilistic embeddings and temporal uncertainty methods under compatible outcomes and information sets. Match architecture, training budget, candidates, and tuning access. A deterministic KG benchmark without confidence labels requires an explicit adaptation before it can be described as an implementation of UKGE.

Report AUROC and error-class AP with prevalence and the stated threshold-grouped convention. ECE and Brier score require probabilities for a named outcome; fit an estimated error probability on development data and evaluate it against $\ell_i$. An uncalibrated novelty score is not a probability. Fix ECE binning in advance and include reliability plots.

Use several training seeds and paired uncertainty estimates for method differences. Query dependence induced by shared entities, relation groups, and time must be respected; use a justified cluster or temporal block resampling scheme and report sensitivity. There is no universal graph bootstrap that automatically resolves all dependence.

\section{Additional empirical comparisons}\label{app:empirical-detail}
These tables preserve the complete original baseline comparisons and the stronger controls on the exploratory pooled population. They use the outcome and fitting contracts in the main text; all means and ranges are descriptive over three optimization seeds.

\subsection{Original-predictor support strata and relation composition}\label{app:original-strata}
\paragraph{Ranking within baseline-confidence strata.}
A diagnostic specified while training was in progress forms quintile edges from the frozen baseline error probability on the development-selection partition. Within each resulting test bin, every score retains $\lfloor0.85n_b\rfloor$ of its $n_b$ queries, and retained errors are pooled across bins. Base + binary LR reduces this pooled risk relative to Base LR by a mean 0.64 percentage points on FB15k-237 (0.51--0.82) and 2.39 points on WN18RR (2.06--2.72). The exploratory role-count reductions are 1.09 and 2.76 points, respectively. The strata change the selection rule and hence the estimand; these values are not additional estimates of the primary global-selection contrast. They describe ranking within coarse baseline-confidence levels, not a conditional-independence or causal test. Shared entities and the same small set of training seeds continue to limit inference.

\paragraph{Directional support and interpretation.}
An exploratory diagnostic chosen after inspecting the primary results separates the two cases assigned the same scalar score $U_C=1/2$. On FB15k-237, mean error is 99.77\% when only predicted-tail support is missing but 64.48\% when only known-head support is missing. A missing-tail indicator has mean error AUROC 0.755, versus 0.504 for a missing-head indicator. The corresponding WN18RR stratum errors are 49.87\% and 67.44\%, with AUROCs 0.632 and 0.703. Thus the useful direction differs between datasets, and equal scalar coverage need not imply similar empirical risk. These post-result comparisons are descriptive failure-case analysis, not confirmatory endpoints. Table~\ref{tab:directional-strata} gives all stratum denominators and the averaging convention.

\begin{table}[ht]
\centering\small
\begin{tabular}{@{}llrrrr@{}}\toprule
Dataset & $(b_H,b_T)$ & Seed 17: $e/n$ & Seed 29: $e/n$ & Seed 43: $e/n$ & Mean error (\%)\\\midrule
FB15k-237 & 0,0 & 3,123/3,695 & 3,157/3,777 & 3,183/3,793 & 84.01 \\
FB15k-237 & 0,1 & 1,962/1,964 & 1,876/1,882 & 1,861/1,866 & 99.77 \\
FB15k-237 & 1,0 & 711/1,102 & 753/1,156 & 789/1,237 & 64.48 \\
FB15k-237 & 1,1 & 3,480/3,540 & 3,415/3,486 & 3,330/3,405 & 98.02 \\
WN18RR & 0,0 & 188/646 & 183/646 & 191/647 & 28.98 \\
WN18RR & 0,1 & 186/380 & 192/380 & 190/379 & 49.87 \\
WN18RR & 1,0 & 536/798 & 582/842 & 513/777 & 67.44 \\
WN18RR & 1,1 & 751/865 & 708/821 & 769/886 & 86.62 \\
\bottomrule
\end{tabular}
\caption{DistMult directional strata under the original candidate rule. Each cell gives errors/queries; the last column is the unweighted mean of the three within-seed error fractions. Predicted-tail membership can vary by seed. Repeated queries across seeds are repeated model evaluations, not independent observations.}\label{tab:directional-strata}
\end{table}

Role coverage can encode whether an entity has previously occupied a relation's head or tail role. The sets defining $c_H$ and $c_T$ are exactly the observed domain/range sets used in the local closed-world approximation of Krompa\ss{} et al.\ \cite{typeconstraints}; the indicators themselves are established features. Here they are used for post-prediction error ranking rather than as constraints in embedding training. DistMult's symmetry between head and tail makes this a particularly relevant competing explanation: post-hoc directional features may identify prediction failures caused by the base architecture or its candidate set. Their value here does not establish a newly identified uncertainty component or an advantage for a probabilistic embedding. The candidate-support diagnostic in the main text probes this explanation; the matched-budget ComplEx check removes mandatory symmetry as an exclusive explanation, but does not establish relevance to a competitive predictor.

\paragraph{Relation-composition sensitivity.}
We also omit each relation in turn from the saved test ledgers, keeping every fitted score fixed and recomputing 85\% global batch selection on the remaining queries. On FB15k-237, the coverage-minus-baseline risk difference remains negative for every omission and all six original predictors. On WN18RR, omitting \texttt{\_synset\_domain\_topic\_of} removes 109 of 2,689 queries and changes the three DistMult differences to $0$, $-0.1824$, and $+0.0456$ percentage points: the mean is $-0.0456$, compared with $-0.2772$ originally. The corresponding ComplEx differences remain $-0.2736$, $-0.3648$, and $-0.3192$ points, mean $-0.3192$. Thus consistency across seeds does not ensure invariance to relation composition. These are altered finite-population rankings, not effects within the omitted relation, independent-cluster inference, or causal estimates.

\subsection{Support strata after reciprocal continued training}
\begin{table}[ht]
\centering\small
\begin{tabular}{@{}llrrrr@{}}\toprule
Dataset & $(b_H,b_T)$ & Seed 17: $e/n$ & Seed 29: $e/n$ & Seed 43: $e/n$ & Mean error (\%)\\\midrule
FB15k-237 & 0,0 & 3,737/5,271 & 3,747/5,266 & 3,683/5,239 & 70.78 \\
FB15k-237 & 0,1 & 382/388 & 389/393 & 415/420 & 98.75 \\
FB15k-237 & 1,0 & 1,239/4,181 & 1,178/4,154 & 1,214/4,152 & 29.08 \\
FB15k-237 & 1,1 & 388/461 & 414/488 & 410/490 & 84.22 \\
WN18RR & 0,0 & 188/689 & 213/715 & 210/710 & 28.88 \\
WN18RR & 0,1 & 135/337 & 108/311 & 111/316 & 36.64 \\
WN18RR & 1,0 & 913/1,300 & 1,002/1,381 & 992/1,387 & 71.44 \\
WN18RR & 1,1 & 246/363 & 165/282 & 158/276 & 61.18 \\
\bottomrule\end{tabular}
\caption{Selected reciprocal ComplEx checkpoints under the original candidate policy. Entries are errors/queries. The final column averages the three within-seed rates equally. Repeated queries and warm-start parent/child runs are not independent observations.}\label{tab:reciprocal-strata}
\end{table}

\begin{table}[t]
\centering
\small
\begin{tabular}{lrr}
\toprule
Score & FB15k-237 & WN18RR \\
\midrule
Score margin & 88.58 [87.96, 89.31] & 55.74 [55.71, 55.75] \\
Entity frequency & 89.18 [88.77, 89.64] & 57.07 [56.96, 57.23] \\
Role coverage & 87.93 [87.39, 88.59] & 57.46 [57.34, 57.62] \\
Fixed mixture & 88.20 [87.69, 88.82] & 57.19 [57.06, 57.40] \\
Selected mixture & 87.93 [87.39, 88.59] & 56.74 [56.68, 56.87] \\
Confidence logistic & 88.55 [88.14, 89.11] & 56.00 [55.71, 56.46] \\
Entity-count logistic & 88.17 [87.54, 88.79] & 56.19 [55.89, 56.59] \\
Coverage logistic & 87.64 [87.06, 88.35] & 55.92 [55.71, 56.15] \\
Role-count logistic & 87.58 [87.00, 88.29] & 55.10 [55.01, 55.19] \\
\midrule
Full-coverage accuracy & 10.56 [9.95, 11.05] & 38.16 [38.08, 38.23] \\
Primary $\Delta$ risk (pp) & -0.54 [-0.69, -0.45] & -0.28 [-0.44, -0.18] \\
\bottomrule
\end{tabular}
\caption{New DistMult experiments. Rows of scores report observed-target error risk (percent) at 85\% batch coverage; smaller is better. Entries are seed means [minimum, maximum]. The primary contrast is coverage logistic minus entity-count logistic (percentage points); negative values favor coverage. Full-coverage accuracy is listed separately. Ranges describe model-seed variability and are not confidence intervals. FB15k-237: $n=10,301$, 3 seeds; WN18RR: $n=2,689$, 3 seeds.}
\label{tab:from-scratch-results}
\end{table}

\begin{table}[t]
\centering
\small
\begin{tabular}{lrr}
\toprule
Score & FB15k-237 & WN18RR \\
\midrule
Score margin & 89.16 [88.95, 89.45] & 56.63 [56.15, 56.94] \\
Entity frequency & 89.40 [89.01, 89.78] & 57.94 [57.49, 58.25] \\
Role coverage & 88.44 [88.15, 88.84] & 58.20 [57.86, 58.48] \\
Fixed mixture & 88.64 [88.29, 89.05] & 58.09 [57.79, 58.42] \\
Selected mixture & 88.44 [88.15, 88.84] & 57.81 [57.49, 58.05] \\
Confidence logistic & 89.08 [88.83, 89.39] & 56.78 [56.28, 57.07] \\
Entity-count logistic & 88.70 [88.50, 89.07] & 56.79 [56.11, 57.16] \\
Coverage logistic & 88.23 [87.94, 88.62] & 56.44 [55.84, 56.76] \\
Role-count logistic & 88.18 [87.90, 88.57] & 55.90 [55.45, 56.24] \\
\midrule
Full-coverage accuracy & 10.05 [9.72, 10.28] & 37.47 [37.19, 37.86] \\
Primary $\Delta$ risk (pp) & -0.47 [-0.56, -0.41] & -0.35 [-0.44, -0.26] \\
\bottomrule
\end{tabular}
\caption{Exploratory ComplEx model sensitivity after the DistMult results. Thirty epochs and 32 complex components match the original epoch and parameter budgets. Rows of scores report observed-target error risk (percent) at 85\% batch coverage; smaller is better. Entries are seed means [minimum, maximum]. The primary contrast is coverage logistic minus entity-count logistic (percentage points); negative values favor coverage. Full-coverage accuracy is listed separately. Ranges describe model-seed variability and are not confidence intervals. FB15k-237: $n=10,301$, 3 seeds; WN18RR: $n=2,689$, 3 seeds.}
\label{tab:complex-results}
\end{table}

\begin{table}[t]
\centering\small
\begin{tabular}{lllrrr}\toprule
Predictor & Dataset & Error head & Risk85 (\%) & AURC & Brier\\\midrule
DistMult & FB15k-237 & Base LR & 86.046 & 0.75820 & 0.09844\\
DistMult & FB15k-237 & Base + binary LR & 84.977 & 0.70033 & 0.08199\\
DistMult & FB15k-237 & Base + counts LR & 84.900 & 0.65979 & 0.06346\\
DistMult & FB15k-237 & Base + counts + binary LR & 84.910 & 0.65611 & 0.06077\\
DistMult & FB15k-237 & Base HGB & 84.973 & 0.69538 & 0.07959\\
DistMult & FB15k-237 & Base + binary HGB & 84.910 & 0.66276 & 0.06550\\
DistMult & FB15k-237 & Base + counts HGB & 84.900 & 0.64576 & 0.05403\\
\addlinespace
DistMult & WN18RR & Base LR & 58.451 & 0.35833 & 0.12216\\
DistMult & WN18RR & Base + binary LR & 58.260 & 0.32709 & 0.10809\\
DistMult & WN18RR & Base + counts LR & 57.317 & 0.33140 & 0.10327\\
DistMult & WN18RR & Base + counts + binary LR & 57.317 & 0.31320 & 0.09186\\
DistMult & WN18RR & Base HGB & 57.435 & 0.27430 & 0.02155\\
DistMult & WN18RR & Base + binary HGB & 57.317 & 0.27242 & 0.01666\\
DistMult & WN18RR & Base + counts HGB & 57.317 & 0.27133 & 0.01370\\
\addlinespace
ComplEx & FB15k-237 & Base LR & 86.557 & 0.77261 & 0.09449\\
ComplEx & FB15k-237 & Base + binary LR & 85.935 & 0.71143 & 0.07734\\
ComplEx & FB15k-237 & Base + counts LR & 85.788 & 0.67488 & 0.06117\\
ComplEx & FB15k-237 & Base + counts + binary LR & 85.791 & 0.67162 & 0.05901\\
ComplEx & FB15k-237 & Base HGB & 85.872 & 0.71384 & 0.07724\\
ComplEx & FB15k-237 & Base + binary HGB & 85.788 & 0.68062 & 0.06406\\
ComplEx & FB15k-237 & Base + counts HGB & 85.788 & 0.66296 & 0.05317\\
\addlinespace
ComplEx & WN18RR & Base LR & 59.025 & 0.29478 & 0.05164\\
ComplEx & WN18RR & Base + binary LR & 58.569 & 0.29127 & 0.04573\\
ComplEx & WN18RR & Base + counts LR & 58.024 & 0.28557 & 0.04038\\
ComplEx & WN18RR & Base + counts + binary LR & 58.009 & 0.28334 & 0.03588\\
ComplEx & WN18RR & Base HGB & 58.098 & 0.28105 & 0.02271\\
ComplEx & WN18RR & Base + binary HGB & 58.009 & 0.27858 & 0.01791\\
ComplEx & WN18RR & Base + counts HGB & 58.039 & 0.27744 & 0.01422\\
\addlinespace
\bottomrule\end{tabular}
\caption{Exploratory stronger scorer controls on the grouped-pool population. All heads use only grouped development-fit outcomes. Base is confidence and entity/relation frequencies; counts adds continuous directional role counts. HGB hyperparameters are identical across feature sets. Means over three seeds are descriptive.}
\label{tab:grouped-score-controls}
\end{table}

\begin{table}[p]
\centering\scriptsize\setlength{\tabcolsep}{3pt}
\begin{tabularx}{\linewidth}{@{}Xrrrr@{}}\toprule
Fixed task & Base & Counts & Relation ID & Relation ID + counts\\\midrule
Ensemble ComplEx / FB15k-237 & 0.729664 / 0.084995 & 0.659133 / 0.054298 & 0.691318 / 0.068342 & 0.651035 / 0.047804\\
Ensemble ComplEx / WN18RR & 0.241836 / 0.024071 & 0.239304 / 0.017701 & 0.241099 / 0.024084 & 0.238684 / 0.013525\\
Ensemble DistMult / FB15k-237 & 0.717466 / 0.088317 & 0.645216 / 0.055849 & 0.674492 / 0.069531 & 0.633191 / 0.046926\\
Ensemble DistMult / WN18RR & 0.271084 / 0.092613 & 0.258631 / 0.073060 & 0.239126 / 0.032299 & 0.234904 / 0.017832\\
Ensemble ReciprocalComplEx / FB15k-237 & 0.228582 / 0.128356 & 0.219667 / 0.117010 & 0.217581 / 0.114699 & 0.216329 / 0.113227\\
Ensemble ReciprocalComplEx / WN18RR & 0.193718 / 0.039378 & 0.192959 / 0.038769 & 0.191109 / 0.038256 & 0.189965 / 0.037613\\
CoDEx reciprocal / CoDEx-S / seed 17 & 0.326474 / 0.197149 & 0.321050 / 0.193951 & 0.316948 / 0.191265 & 0.312564 / 0.188043\\
CoDEx reciprocal / CoDEx-S / seed 29 & 0.320963 / 0.189586 & 0.314462 / 0.185179 & 0.315409 / 0.187588 & 0.309169 / 0.183435\\
CoDEx reciprocal / CoDEx-S / seed 43 & 0.334027 / 0.198574 & 0.328581 / 0.195053 & 0.334268 / 0.198158 & 0.329543 / 0.195415\\
CoDEx sampled / CoDEx-S / seed 17 & 0.736701 / 0.120562 & 0.623799 / 0.081676 & 0.714065 / 0.114610 & 0.619629 / 0.079328\\
CoDEx sampled / CoDEx-S / seed 29 & 0.750555 / 0.111668 & 0.634032 / 0.067833 & 0.726105 / 0.107114 & 0.628329 / 0.072048\\
CoDEx sampled / CoDEx-S / seed 43 & 0.769650 / 0.105754 & 0.655817 / 0.067522 & 0.738037 / 0.097452 & 0.651535 / 0.064709\\
Reciprocal / FB15k-237 / seed 17 & 0.261423 / 0.133850 & 0.247565 / 0.118245 & 0.246865 / 0.117336 & 0.242625 / 0.111848\\
Reciprocal / FB15k-237 / seed 29 & 0.257487 / 0.131181 & 0.243234 / 0.115126 & 0.243499 / 0.114969 & 0.238512 / 0.109055\\
Reciprocal / FB15k-237 / seed 43 & 0.258494 / 0.133816 & 0.244284 / 0.117444 & 0.244213 / 0.117260 & 0.239310 / 0.111159\\
Reciprocal / WN18RR / seed 17 & 0.207665 / 0.055807 & 0.205221 / 0.051851 & 0.203045 / 0.046004 & 0.200089 / 0.042671\\
Reciprocal / WN18RR / seed 29 & 0.210984 / 0.056572 & 0.209631 / 0.054359 & 0.205486 / 0.043916 & 0.203661 / 0.041468\\
Reciprocal / WN18RR / seed 43 & 0.200814 / 0.046369 & 0.199586 / 0.042929 & 0.198673 / 0.043015 & 0.196741 / 0.040251\\
\bottomrule\end{tabularx}
\caption{AURC / Brier. All four LR feature variants use the same C grid and development-AURC selection. Base includes confidence/frequencies and, for ensembles, explicit disagreement. Original two-graph controls are exploratory; CoDEx-S heads were fitted and selected before its first test evaluation. Seed means are descriptive; each ensemble row is one fixed ensemble, not a replicated estimate.}\label{tab:relation-identity-scores}
\end{table}

\begin{table}[p]
\centering\scriptsize\setlength{\tabcolsep}{3pt}
\begin{tabularx}{\linewidth}{@{}Xrrrr@{}}\toprule
Fixed task & Base & Counts & Relation ID & Relation ID + counts\\\midrule
Ensemble ComplEx / FB15k-237 & 78.641 / 85.791 & 75.534 / 85.574 & 77.883 / 85.905 & 75.612 / 85.574\\
Ensemble ComplEx / WN18RR & 22.396 / 53.260 & 20.536 / 53.217 & 22.173 / 53.348 & 20.610 / 53.217\\
Ensemble DistMult / FB15k-237 & 77.650 / 85.094 & 74.155 / 84.763 & 76.058 / 84.969 & 74.136 / 84.763\\
Ensemble DistMult / WN18RR & 25.000 / 52.691 & 21.503 / 52.560 & 22.173 / 52.648 & 19.866 / 52.516\\
Ensemble ReciprocalComplEx / FB15k-237 & 21.553 / 45.208 & 20.078 / 44.900 & 19.748 / 44.991 & 19.786 / 44.866\\
Ensemble ReciprocalComplEx / WN18RR & 11.161 / 46.346 & 11.235 / 46.083 & 11.086 / 46.083 & 11.161 / 45.864\\
CoDEx reciprocal / CoDEx-S / seed 17 & 35.870 / 49.201 & 34.783 / 49.201 & 33.696 / 48.882 & 33.333 / 48.243\\
CoDEx reciprocal / CoDEx-S / seed 29 & 35.326 / 50.266 & 34.601 / 49.947 & 33.877 / 50.266 & 33.696 / 49.521\\
CoDEx reciprocal / CoDEx-S / seed 43 & 35.870 / 48.775 & 34.601 / 48.456 & 35.688 / 48.562 & 35.145 / 48.349\\
CoDEx sampled / CoDEx-S / seed 17 & 77.717 / 82.854 & 68.841 / 81.683 & 76.087 / 82.002 & 68.841 / 81.683\\
CoDEx sampled / CoDEx-S / seed 29 & 77.899 / 84.026 & 71.377 / 83.174 & 76.812 / 83.600 & 71.377 / 83.174\\
CoDEx sampled / CoDEx-S / seed 43 & 80.616 / 85.623 & 73.551 / 84.452 & 78.804 / 85.091 & 73.551 / 84.452\\
Reciprocal / FB15k-237 / seed 17 & 27.068 / 48.978 & 24.447 / 48.384 & 24.078 / 48.692 & 23.204 / 48.361\\
Reciprocal / FB15k-237 / seed 29 & 26.757 / 48.681 & 23.709 / 48.121 & 23.573 / 48.384 & 22.466 / 47.995\\
Reciprocal / FB15k-237 / seed 43 & 26.718 / 48.658 & 23.961 / 48.155 & 23.883 / 48.258 & 22.835 / 48.075\\
Reciprocal / WN18RR / seed 17 & 15.104 / 47.877 & 14.509 / 47.440 & 14.881 / 47.440 & 14.137 / 47.265\\
Reciprocal / WN18RR / seed 29 & 15.923 / 48.053 & 15.625 / 47.877 & 15.327 / 47.921 & 14.658 / 47.659\\
Reciprocal / WN18RR / seed 43 & 13.616 / 47.002 & 12.946 / 46.915 & 13.318 / 46.915 & 12.649 / 46.915\\
\bottomrule\end{tabularx}
\caption{Risk50 / Risk85 (\%). All four LR feature variants use the same C grid and development-AURC selection. Base includes confidence/frequencies and, for ensembles, explicit disagreement. Original two-graph controls are exploratory; CoDEx-S heads were fitted and selected before its first test evaluation. Seed means are descriptive; each ensemble row is one fixed ensemble, not a replicated estimate.}\label{tab:relation-identity-risks}
\end{table}

\begin{figure}[t]
\centering
\includegraphics[width=\linewidth]{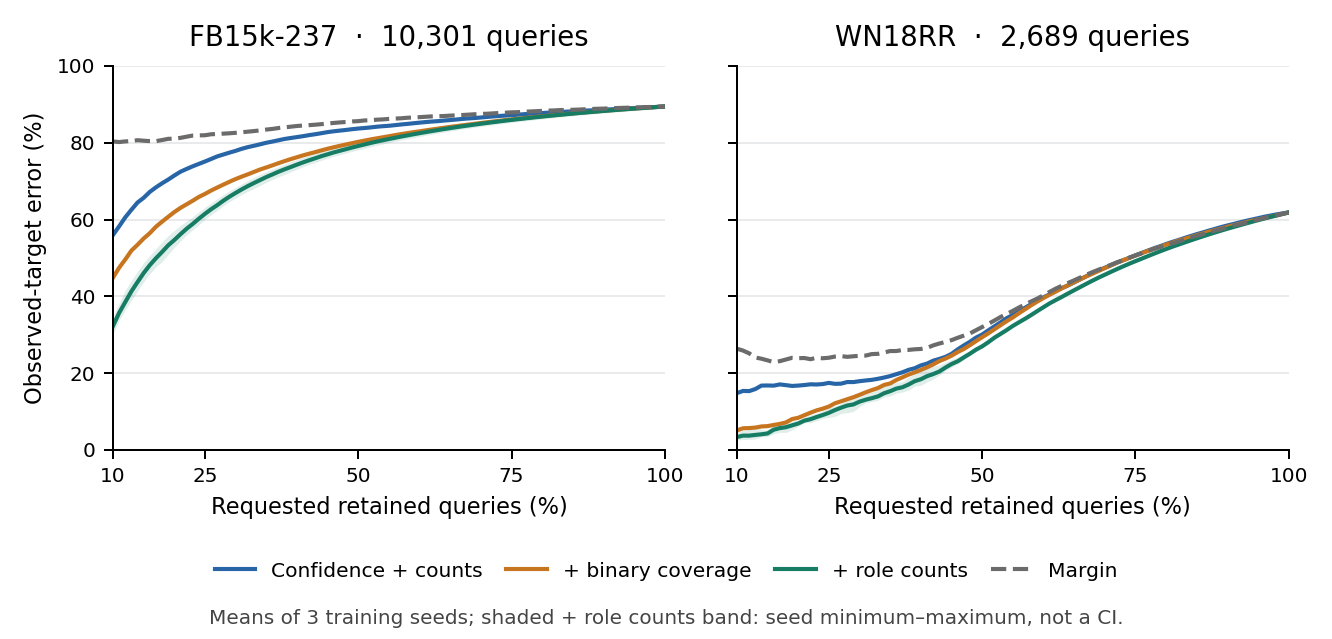}
\caption{Original DistMult nine-score study: observed-target error risk against requested batch coverage. Lines are means over three training seeds. The role-count-logistic shaded band is the seed minimum--maximum, not a confidence interval. Score ties use analytic random-order expectation. Stronger count and HGB comparisons and their hindsight floors appear in the main text.}\label{fig:benchmark-risk}
\end{figure}

\clearpage
\section{Additional metric and inference conventions}\label{app:metrics}
\paragraph{Average precision and paired inference.}
For error detection we implement noninterpolated average precision (AP). Order distinct uncertainty levels from largest to smallest; if block $j$ contains $e_j$ errors, let $E_j$ and $N_j$ be cumulative errors and queries. With $E=\sum_j e_j>0$,
\[
 \operatorname{AP}=\sum_j\frac{e_j}{E}\frac{E_j}{N_j}.
\]
This convention includes a whole tied block at each threshold and differs from both trapezoidal PR area and randomized within-tie rankings. Report prevalence alongside AP; a constant score has AP equal to prevalence. With no errors AP is undefined here; with all errors it is one while AUROC remains undefined.

Paired ledgers must match query IDs, predictions, correctness, predictor IDs and candidate-set IDs. Optional percentile intervals resample whole declared independent clusters, using identical draws for both methods and recomputing each risk curve. The estimand remains query-weighted, preserving unequal cluster sizes. These intervals condition on frozen fitted models and tuning choices, and are pointwise. They exclude retraining variability and do not provide simultaneous coverage over scores, datasets or thresholds. Shared entities can induce dependence across relation clusters; merely supplying relation IDs does not justify independent-cluster inference. With no defensible sampling units, report descriptive differences and leave inferential claims open.

\end{document}